%% file: _main.tex
\input{_constants}

\arxiv
\pdfoutput=1
\documentclass[10pt,twocolumn,letterpaper]{article}

\input{cvpr_header}

\usepackage[accsupp]{axessibility} % Improves PDF readability for those with disabilities.

\begin{document}
%% TITLE
\title{\emph{RaBit}: Pa\textit{ra}metric Modeling of 3D \textit{Bi}ped Car\textit{t}oon Characters \\ with a Topological-consistent Dataset}
\author{\authorBlock}
%\maketitle
%%

\twocolumn[{%
  \renewcommand\twocolumn[1][]{#1}%
\maketitle
\begin{center}
  \newcommand{\teaserwidth}{\textwidth}
  \vspace{-0.15in}
  \centerline{
    \includegraphics[width=.95\teaserwidth]{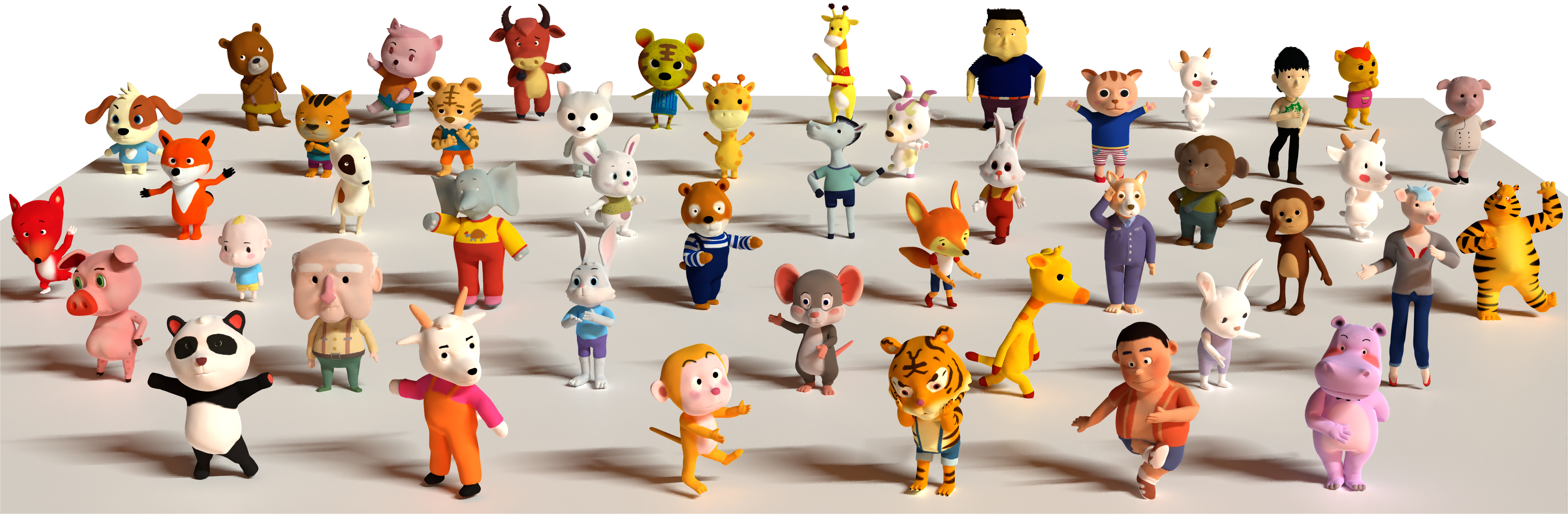}
    }
    % \vspace{-10pt}
    \captionof{figure}{
     We present \textit{\textbf{3DBiCar}}, the first large-scale repository of 3D biped cartoon characters. It contains 1,500 topologically consistent, textured, and skinned 3D high-quality meshes manually created by professional artists, covering 15 species. Further, we propose \textit{RaBit}, the first cartoon character parametric model simultaneously parameterizing shape, pose, and texture. %With \textit{3DBiCar} and \textit{RaBit}, we conduct various applications to demonstrate their promising potential for efficient biped character digitization.
    }
  %\vspace{-0.05in}
  \label{fig_teaser}
 \end{center}%
    }]
\authorFootnote

\input{paper/00_abstract}
\input{paper/01_intro}

\input{paper/02_related}
\input{paper/03_dataset_remake}
\input{paper/04_method}
\input{paper/05_application}
% % \input{paper/06_experiment}
\input{paper/07_conclusion}
{\small
\bibliographystyle{ieee_fullname}
\bibliography{11_references}
}

\ifarxiv \clearpage \input{12_appendix} \fi

\end{document}

%% file: _constants.tex
\def\cvprPaperID{275}
\def\confName{CVPR}
\def\confYear{2023}

\def\authorBlock{
    Zhongjin Luo${}^{1}$\thanks{Equal contribution} \qquad
    Shengcai Cai${}^{1,3}$\footnotemark[1] \qquad
    Jinguo Dong${}^{1}$ \qquad
    Ruibo Ming${}^{1,4}$ \qquad \\
    Liangdong Qiu${}^{2,1}$ \qquad
    Xiaohang Zhan${}^{3}$ \qquad
    Xiaoguang Han${}^{1,2}$\thanks{Corresponding Author} \qquad
    \\
    \\
    ${}^{1}$SSE, CUHKSZ  \quad ${}^{2}$FNii, CUHKSZ \quad    ${}^{3}$ Huawei Technologies Co., Ltd. \quad ${}^{4}$Tsinghua University
}

\def\authorFootnote{
\renewcommand{\thefootnote}{\fnsymbol{footnote}}
\footnotetext[1]{Z. Luo and S. Cai contribute equally.} %对应脚注[1]
\footnotetext[2]{Corresponding author.} %对应脚注[2]
}
\newif\ifreview 
\newif\ifarxiv \newcommand{\arxiv}{\arxivtrue}
\newif\ifcamera 
\newif\ifrebuttal

%% file: cvpr_header.tex
\ifreview \usepackage[review]{cvpr} \fi
\ifarxiv \usepackage[pagenumbers]{cvpr} \fi
\ifrebuttal \usepackage[rebuttal]{cvpr} \fi
\ifcamera \usepackage{cvpr} \fi

\usepackage{graphicx}
\usepackage{amsmath}
\usepackage{amssymb}
\usepackage{booktabs}

\input{_macros}  % Add packages to _macros.tex

\usepackage{xr-hyper}

\makeatletter
\newcommand*{\addFileDependency}[1]{
  \typeout{(#1)}
  \@addtofilelist{#1}
  \IfFileExists{#1}{}{\typeout{No file #1.}}
}

\makeatother

\usepackage[pagebackref,breaklinks,colorlinks]{hyperref}
\usepackage[capitalize]{cleveref}
\crefname{section}{Sec.}{Secs.}
\crefname{table}{Table}{Tables}
\crefname{figure}{Fig.}{Figs.}

%% file: _macros.tex
\usepackage{times}
\usepackage{microtype}
\usepackage{epsfig}
\usepackage[table,xcdraw]{xcolor}
\usepackage{caption}
\usepackage{float}
\usepackage{placeins}
\usepackage{color, colortbl}
\usepackage{stfloats}
\usepackage{enumitem}
\usepackage{tabularx}
\usepackage{xstring}
\usepackage{multirow}
\usepackage{xspace}
\usepackage{url}
\usepackage{subcaption}
\usepackage{xcolor}
\usepackage[hang,flushmargin]{footmisc}

\ifcamera \usepackage[accsupp]{axessibility} \fi

% ...

\ifarxiv  \fi

\newcommand{\R}[1]{{%
    \textbf{%
        \ifstrequal{#1}{1}{\textcolor{red}{R#1}}{%
        \ifstrequal{#1}{2}{\textcolor{blue}{R#1}}{%
        \ifstrequal{#1}{3}{\textcolor{magenta}{R#1}}{%
        \ifstrequal{#1}{4}{\textcolor{teal}{R#1}}{%
                           \textcolor{cyan}{R#1}%
        }}}}%
    }%
}}

%% file: paper/00_abstract.tex
\begin{abstract}
 Assisting people in efficiently producing visually plausible 3D characters has always been a fundamental research topic in computer vision and computer graphics. Recent learning-based approaches have achieved unprecedented accuracy and efficiency in the area of 3D real human digitization. However, none of the prior works focus on modeling 3D biped cartoon characters, which are also in great demand in gaming and filming. In this paper, we introduce \textit{3DBiCar}, the first large-scale dataset of 3D biped cartoon characters, and \textit{RaBit}, the corresponding parametric model. Our dataset contains 1,500 topologically consistent high-quality 3D textured models which are manually crafted by professional artists. Built upon the data, \textit{RaBit} is thus designed with a SMPL-like linear blend shape model and a StyleGAN-based neural UV-texture generator, simultaneously expressing the shape, pose, and texture. To demonstrate the practicality of \textit{3DBiCar} and \textit{RaBit}, various applications are conducted, including single-view reconstruction, sketch-based modeling, and 3D cartoon animation. For the single-view reconstruction setting, we find a straightforward global mapping from input images to the output UV-based texture maps tends to lose detailed appearances of some local parts (e.g., nose, ears). Thus, a part-sensitive texture reasoner is adopted to make all important local areas perceived. Experiments further demonstrate the effectiveness of our method both qualitatively and quantitatively. \textit{3DBiCar} and \textit{RaBit} are available at \href{https://gaplab.cuhk.edu.cn/projects/RaBit/}{gaplab.cuhk.edu.cn/projects/RaBit}.
   
   %We also propose to utilize cumulative local UV mappings to enhance local details in the single-view reconstruction. Our various application results show the promising potential of \textit{3DBiCar}, the outstanding capability of \textit{RaBit}, and the superb performance of our single-view reconstruction method. 
   
   %We will make \textit{3DBiCar} and \textit{RaBit} publicly available upon publication.
\end{abstract}

%% file: paper/01_intro.tex
\vspace{-0.5cm}
\section{Introduction}
\label{sec:intro}

With the rapid development of digitization, creating high-quality 3D articulated characters is highly demanded in game platforms, film industries, and metaverse scenarios. However, even for expert artists, creating a 3D character is labor-intensive and time-consuming. Therefore, reducing the cost of producing visually plausible 3D characters is essential in the field of computer vision and graphics.

Recently, researchers have made great progress in digitizing realistic human characters. The emergence and popularity of various 3D sensing devices make capturing 3D data from the real world convenient, prompting a growing number of 3D real-people scanned datasets~\cite{cao2013facewarehouse,yang2020faceScape,robinette2002civilian,anguelov2005scape,faust:CVPR:2014,yang2014spring,saint2018-3dBodyTex,Zheng2019DeepHuman}. Based on these large-scale datasets, several powerful parametric models~\cite{SMPL:2015,SMPL-X:2019,blanz1999morphable,cao2013facewarehouse} have been developed to facilitate the reconstruction and analysis of human shapes, actions, and interactions. With the help of parametric models, deep learning techniques have shown the potential to efficiently infer accurate 3D digital humans from single-view images~\cite{SMPL-X:2019,kanazawa2018end} or even sparse sketches~\cite{han2017deepsketch2face,brodt2022sketch2pose,unlu2022interactive}. Most recently, there are some works~\cite{qiu20213dcaricshop,luo2021simpmodeling} that devote to exploring the intelligent generation of cartoon-like character heads. However,
none of the prior works focuses on the modeling of 3D full-body biped cartoon characters, which are also in great demand in the area of gaming (e.g., Animal Crossing), filming (e.g., Zootopia), and virtualizing (e.g., Metaverse). In this work, we raise a new problem to the community: \emph{How to quickly produce 3D biped cartoon characters from easy-to-obtain inputs (e.g., a single image)?}

%How to quickly produce 3D biped cartoon characters from simple input, e.g., a single image or sparse strokes, still be an open problem that needs to be solved urgently.

%Thus, creating 3D biped cartoon characters from simple input, e.g., a single image or sparse strokes, is of crucial topic for 3D character digitization.

Revisiting the road map of realistic human digitization, the first step to tackling the above problem is building a high-quality 3D biped cartoon characters dataset. We thus introduce \textit{3DBiCar}, the first large-scale publicly available 3D biped cartoon character dataset following three criteria: \textit{1) \textbf{Diversity.}} \textit{3DBiCar} spans a wide range of 3D biped cartoon characters, containing 1,500 high-quality 3D models covering 15 species, as shown in the Fig.~\ref{fig_datainfo}. \textit{2) \textbf{Richness.}} Each model in \textit{3DBiCar} owns not only a detailed shape but also a texture UV-map, which are matched with a reference image. Additionally, each character is attached with two models, one with T-pose and another with the reference pose. \textit{3) \textbf{Topological-consistency.}} Each 3D model is created by carefully deforming a pre-defined template mesh. All 3D characters in \textit{3DBiCar} are unified in topology, paving the way to learn a skinned parametric model. Fig.~\ref{fig_teaser} shows some representative models of the proposed dataset.
%Therefore, \textit{3DBiCar} provides the shape, pose, texture map, and the corresponding 2D reference image for each model simultaneously, which could be directly applied to several vital tasks in visual computing such as single-view reconstruction, pose tracking, and texture synthesis.

Based on \textit{3DBiCar}, we further propose a generative model, dubbed \textit{RaBit}, for 3D biped cartoon character generation. It combines a linear blend shape model with a neural texture generator and simultaneously parameterizes the shape, pose, and texture to a low-dimensional parametric space. For shape and pose modeling, numerous methods have shown principal component analysis's (PCA's) advantage %super ability 
in building decent statistical shape models~\cite{SMPL-X:2019, blanz1999morphable, cao2013facewarehouse, FLAME:SiggraphAsia2017}. Inspired by SMPL~\cite{SMPL:2015}, we utilize the traditional PCA technique to parameterize shape. Due to the variety and complexity of cartoon texture, directly adopting PCA for texture modeling fails to reconstruct details and falls into blurry results. We tackle this problem by introducing a StyleGAN-based generator. 

% Next, with \textit{3DBiCar}, a 3D cartoon character parametric model \textit{BiCar} is further proposed to model the 3D fictitious biped character encoding shape, pose, and texture at the same time for generalized character digitization. \textit{BiCar} is a statistical model for 3D cartoon character generation, which embeds shape, pose, and texture to a low-dimensional parametric space simultaneously. For shape and pose modeling, with pioneering human parameterization~\cite{SMPL:2015}, numerous methods~\cite{MANO:SIGGRAPHASIA:2017, SMPL-X:2019, STAR:2020} have been proposed to learn 3D human model representation from scan data, and these parametric models are widely used for digitizing humans from input images~\cite{kanazawa2018end, kolotouros2019spin, SMPL-X:2019}. However, straightly adapting these blendershape approaches to \textit{3DBiCar} fails to reconstruct high-fidelity models texture from low-dimension parameters due to the diversity and complexity texture of cartoon characters, especially their heads. We tackled this problem by introducing a non-linear generative model parameterization. Current texture modeling benefiting from deep representation learning advances, several works~\cite{bengio2013representation, kingma2013auto, karras2019style, karras2020analyzing} served well in embedding images to a low-dimension latent space. Inspired by image domain transfer methods~\cite{karras2020analyzing, wang2018pix2pixHD, karras2019style}, we encode texture into a latent parametric space and adopt a UV generator to parameterize texture.

To explore the practical usage of \textit{3DBiCar} and \textit{RaBit}, we first conduct the application of single-view reconstruction. Considering prior works for SMPL-based human geometry generation from single-view images\cite{kanazawa2018end,kolotouros2019spin,bogo2016keep}, we build a baseline method with our dataset and the parametric model. We select one regression-based method for pose and shape inference. For texture inference, we find directly applying a global texture-generator tends to make the results lose detailed appearances, especially for some local but important regions (e.g., nose and ears). Thus a part-sensitive reasoner is utilized to deal with different local regions. We term our baseline method for single-view reconstruction as \emph{BiCarNet}. Moreover, two further applications, i.e., sketch-based modeling and 3D character animation, are also explored. Experimental results on these applications demonstrate that it is already able to generate reasonable outputs. %We hope that our work opens a door for researchers to explore bipedal character digitization with the proposed dataset.

%Furthermore, we conduct comprehensive experiments to explore the potential of \textit{3DBiCar} and evaluate the capability of \textit{RaBit}. Specifically,  we implement exhaustive applications, including single-view cartoon character reconstruction, sketch-based character modeling, and 3D cartoon animation. In the single-view reconstruction, we further propose to integrate cumulative local UV mappings to further enhance the representation ability of local details (e.g., nose, ears) of texture. Experimental results on these applications not only demonstrate the usability of \textit{3DBiCar} but also the outstanding capability of \textit{RaBit}. We hope that our work \textit{3DBiCar} and \textit{RaBit} will boost further research on efficient bipedal character digitization. 

% \zj{To summarize, we propose \textit{3DBiCar}, the first large-scale 3D biped cartoon character dataset, which contains 1,500 meticulously crafted, textured 3D models with a consistent mesh topology. Based on this dataset, we build the first 3D full-body cartoon parametric model \textit{RaBit} for biped character modeling. Both \textit{3DBiCar} and \textit{RaBit} will be released to facilitate future research. To demonstrate the promising potential of \textit{3DBiCar} and \textit{RaBit}, we conduct various downstream applications, including single-view reconstruction, sketch-based modeling, and 3D cartoon animation. We hope that our work will contribute to the development of 3D biped cartoon character modeling and inspire future works in this area.}
To summarize, our contributions include:
\begin{itemize}
    \item We introduce \textit{3DBiCar}, the first large-scale 3D biped cartoon character dataset. It contains 1,500 high-quality textured 3D models with a consistent mesh topology.
    
    \item We propose \textit{RaBit}, the first 3D full-body cartoon parametric model for biped character modeling. We will release both \textit{3DBiCar} and \textit{RaBit} for future research.
    
    % \item We carefully designed \emph{BiCarNet}, the first method to reconstruct 3D textured biped cartoon characters from a single-view image. A novel part-sensitive reasoner is invented for detailed texture generation.  
    \item We build \emph{BiCarNet}, the baseline method to reconstruct 3D biped cartoon characters from a single-view image. A part-sensitive reasoner is adopted for detailed texture generation.
    
    %We conduct various applications to demonstrate the promising potential of \textit{3DBiCar} and \textit{RaBit} in single-view cartoon character reconstruction, sketch-based character modeling, and 3D cartoon animation. 

    % \item Two other applications, i.e., sketch-based modeling and 3D character animation, are also successfully conducted. We strongly believe our work opens a door for future research. 
    \item Two other applications, i.e., sketch-based modeling and 3D character animation, are also conducted to demonstrate the promising potential of \textit{3DBiCar} and \textit{RaBit}.
    
    %We propose to adopt additional local UV mappings to further enhance the representation ability of local details in the single-view reconstruction. Quantitative experiments demonstrate the superior of our method.
\end{itemize}

%% file: paper/02_related.tex
\section{Related Work}
\label{sec:related_works}
\noindent
\textbf{3D Character Datasets.} In general, 3D character datasets could be categorized as real-captured and computer-designed datasets. For capturing character data from the real world, the availability of 3D scanning devices has enabled researchers to collect many scanned 3D human-related datasets, mainly focusing on human faces~\cite{brunton2014review} and bodies~\cite{cheng2018parametric}. For human faces,  FaceWarehouse~\cite{cao2013facewarehouse} collects large-scale 3d faces with high diversity in age, ethnicity, and expression. FaceScape\cite{yang2020faceScape} further builds a large-scale detailed 3D face dataset with high resolution in texture and mesh. For human bodies, CAESAR dataset~\cite{robinette2002civilian} opens up the learning of the human body and is widely used for body shape modeling for its shape diversity and satisfying resolution of meshes. Many following works~\cite{anguelov2005scape,faust:CVPR:2014,yang2014spring,saint2018-3dBodyTex,Zheng2019DeepHuman} extend ~\cite{robinette2002civilian} in shape, pose, and texture, on quantity and quality. Although these real-captured datasets are widely used in realistic human digitalization, they are unsuitable for imaginary character generation.

For designing character data with computers, researchers try to perform deformation on real 3D human faces or bodies to construct exaggerated shapes programmatically~\cite{sela2015computational,han2017deepsketch2face,cai2021landmark,wu2018alive}. Their results lack diversity and are far from satisfactory. To address this, 3DCaricShop~\cite{qiu20213dcaricshop} proposes a large-scale 3D exaggerated faces dataset. SimpModeling~\cite{luo2021simpmodeling} constructs a large man-made animalmorphic head dataset. Although using 3DCaricShop and SimpModeling could facilitate the generation of unreal character heads, it still remains a problem to synthesize full-body cartoon characters due to the lack of corresponding body data. In our work, we tackle this problem by introducing a large 3D biped cartoon character dataset, \textit{3DBiCar}. %It contains 1,500 high-quality 3D full-body textured models and spans a wide range of biped cartoon characters.

\noindent
\textbf{Parametric Shape Modeling.} Parametric models of shapes are widely used in 3D digitizations. Blanz \textit{et~al.}~\cite{blanz1999morphable} pioneer parametric modeling using PCA and release a 3D statistical morphable face model (3DMM). Their parameterization models textured faces and provides a set of controls for intuitively manipulating shapes, expressions, and textures. Since then, PCA-based parameterizing has gradually dominated the area of statistical shape modeling over the past decades. Following 3DMM, researchers model the whole head to represent the neck region and 3D head rotations~\cite{cao2013facewarehouse, FLAME:SiggraphAsia2017}. Allen \textit{et al.}~\cite{allen2003space} open up the study of full body parameterization. However, they focus only on body shape and omit the body pose. SCAPE~\cite{anguelov2005scape} represents body shape and pose in terms of triangle deformations, while SMPL~\cite{SMPL:2015} models a whole range of natural shapes and poses based on vertex displacements. SMPL-X~\cite{SMPL-X:2019} integrates SMPL~\cite{SMPL:2015} with  FLAME~\cite{FLAME:SiggraphAsia2017} head model and the MANO~\cite{MANO:SIGGRAPHASIA:2017} hand model for expressive capturing of bodies, hands and faces together. With recent advances in deep learning, researchers turn to explore nonlinear shape models using neural networks~\cite{abrevaya2018multilinear,bagautdinov2018modeling,ranjan2018generating,bouritsas2019neural,xu2020ghum,zhou2020fully}. However, since these non-linear modeling methods are inferior in simplicity, robustness and availability, PCA-based methods remain prevalent in the research community. In this paper, we also adopt PCA into \textit{RaBit} to model the geometry of 3D biped cartoon characters.

Based on the above parametric models, researchers have made remarkable progress in human digitization, such as reconstruction from simple inputs (e.g., a single image or sparse sketches)~\cite{bogo2016keep,kanazawa2018end,pavlakos2018learning,choutas2022accurate,han2017deepsketch2face} and real-time pose retargeting~\cite{choi2021beyond,wang2022live,kocabas2020vibe}. For instance, SMPLify~\cite{bogo2016keep} estimates 3D body shape and pose parameters automatically from 2D joints with multiple ellipsoids. HMR~\cite{kanazawa2018end} proposes an end-to-end framework for reconstructing a full 3D mesh of a human body from a single RGB image. DeepSketch2Face~\cite{han2017deepsketch2face} proposes a sketch-based system for inferring 3D face models from 2D sketches with the help of parametric models and CNN-based deep regression networks. TCMR~\cite{choi2021beyond} presents a temporally consistent mesh recovery system for recovering smooth 3D human motion from monocular videos. To probe the capability of \textit{RaBit} to downstream applications, we conduct various  utilization, such as single-view cartoon character reconstruction, sketch-based character modeling, and 3D cartoon animation. Experimental results demonstrate the practicality of \textit{3DBiCar} and \textit{RaBit}. 

\noindent
\textbf{Parametric Texture Modeling.}
Traditionally, textures are modeled as a linear subspace using the similar idea of body blendshape models. Blanz \etal~\cite{blanz1999morphable} represent the face appearance in per-vertex colors and parameterize texture as a linear model based on PCA. Dai \textit{et~al.}~\cite{dai20173d} store texture information in a UV space where the texture resolution is unconstrained by mesh resolution. Moschoglou \textit{et~al.}~\cite{moschoglou2018multi} formulate a robust matrix factorization problem to learn the parametric representation of facial UV maps from a collection of training textures. However, these linear texture models may lead to a sub-optimal appearance output~\cite{egger2016copula,han2012semiparametric} due to the weak assumption of Gaussian and tend to produce blurry results.

With recent advances in deep learning, researchers turn to utilize deep neural networks to model texture. A number of deep generative models~\cite{li2020learning,slossberg2018high,deng2018uv,gecer2019ganfit,grigorev2021stylepeople,fu2022stylegan,oechsle2019texture,gao2022get3d} have been proposed to parameterize texture into a latent space. For example, GANFIT~\cite{gecer2019ganfit} utilizes GAN-based neural networks to train a generator of facial texture in UV space for 3D face reconstruction. StylePeople~\cite{grigorev2021stylepeople} incorporates neural texture synthesis, mesh rendering, and neural rendering into the joint generation process to train a neural texture generator for the task of single-view human reconstruction. GET3D~\cite{gao2022get3d} introduces a texture-field generative model that directly generates explicit textured 3D meshes, ranging from cars, chairs, animals, motorbikes, and human characters to buildings. These methods have shown the promising capacity of neural generators to represent texture. In our work, we adopt a GAN-based neural texture generator into \textit{RaBit} to provide high-quality texture modeling. %Furthermore, in the task of single-view reconstruction, we design a part-sensitive texture reasoner to make all important local appearances perceived.

%Furthermore, we propose integrating cumulative local UV mappings to enhance the local details in texture inference from a single-view image.

%% file: paper/03_dataset_remake.tex
\section{Dataset}
\label{sec:dataset}

\input{figure/fig_dataset_gallery}
\input{figure/fig_table}
\input{figure/fig_dataset_pipeline}

%Recently, researchers have made significant progress in digitizing realistic and articulated human characters. It is already possible to generate relatively accurate 3D real humans from simple inputs, even a single-view image or sparse strokes. However, no existing works focused on the efficient generation of 3D biped cartoon characters, which a great demand in gaming and filming. 
Considerable progress has been made in digitizing realistic and articulated human characters. However, efficiently creating visually plausible biped cartoon characters remains demanding and challenging, mainly due to the lack of data. In this work, we propose to fill this gap by introducing \textit{3DBiCar}, the first large-scale full-body 3D biped character data. We build \textit{3DBiCar} following three rules:

% The lack of large-scale 3D character datasets in the past has led to the stagnation of this field, so we build \textit{3DBiCar}, a large-scale topologically consistent 3D biped cartoon character dataset. In the following part, We will elaborate on the pipeline to build \textit{3DBiCar}, and for a brief dataset demonstration, please refer to Fig.~\ref{fig_dataset_gallery}. 

% image collection varies
% We start establishing our \textit{3DBiCar} by searching diverse images. We carefully select 1,500  biped character images among 17 species in 4 styles from the Internet and e-books. They are vital references for later 3D model crafting and challenging inputs for possible reconstruction tasks. %precious resource?

\textbf{Diversity.} \textit{3DBiCar} spans a wide range of 3D biped cartoon characters, containing 1,500 high-quality 3D models. First, we carefully collect images of 2D full-body biped cartoon characters with diverse identities, shape, and textural styles from the Internet, resulting in 15 character species and 4 image styles, as shown in Fig.~\ref{fig_datainfo}. Then we recruit six professional artists to create 3D corresponding character models according to the collected reference images. The modeling result is required to be matched with the reference images as much as possible. The representative image-model pairs sampled from our dataset are shown in Fig.~\ref{fig_dataset_gallery}. 

\textbf{Topological-consistency.} 
The key to building a linear parametric shape model is keeping a unified mesh topology. Traditional human parametric models utilize a template mesh to register different human body scans with 3D landmarks to keep topologically uniform. Inspired by this, we first create a template mesh with several 3D colored landmarks as shown in Fig.~\ref{fig_template_design}. All six artists are required to craft 3D models by deforming the above-predefined template under the constraints of these obvious landmarks. We set up a review committee of 10 to check these models based on the landmarks, ensuring the consistency of mesh topology. The landmarks could also be used to compute the position of models' joints for body posing or character animation. The topological consistency of \textit{3DBiCar} paves the way to learn a skinned parametric model, which we will discuss in Sec.~\ref{sec:algorithm}.

\textbf{Richness.} We provide various forms of data for each character. There are not only the 3D shape meshes and UV-space textures carefully crafted by artists but also collected reference images. For each character, artists are asked first to create a T-pose mesh and then deform it to match the reference pose. Furthermore, all the models are rigged and skinned using a predefined skeleton and skinning weight matrix, which enables further animation production for characters. In addition, each character contains two separate eyeballs specifically designed for facial animation. The body mesh of each character comprises 38,726 vertices and 77,448 faces, while each eyeball consists of 1,025 vertices and 2,046 faces.

% The artists are asked to produce the corresponding 3D shape, pose and texture according to the reference images. Therefore, \textit{3DBiCar} provides the shape, pose, texture map, and the corresponding 2D reference image for each model simultaneously, which could be directly applied to several vital tasks in visual computing such as single-view reconstruction, pose tracking, and texture synthesis.

%% file: figure/fig_dataset_gallery.tex
\begin{figure*}[htb]
  \centering
  \includegraphics[width=.98\linewidth]{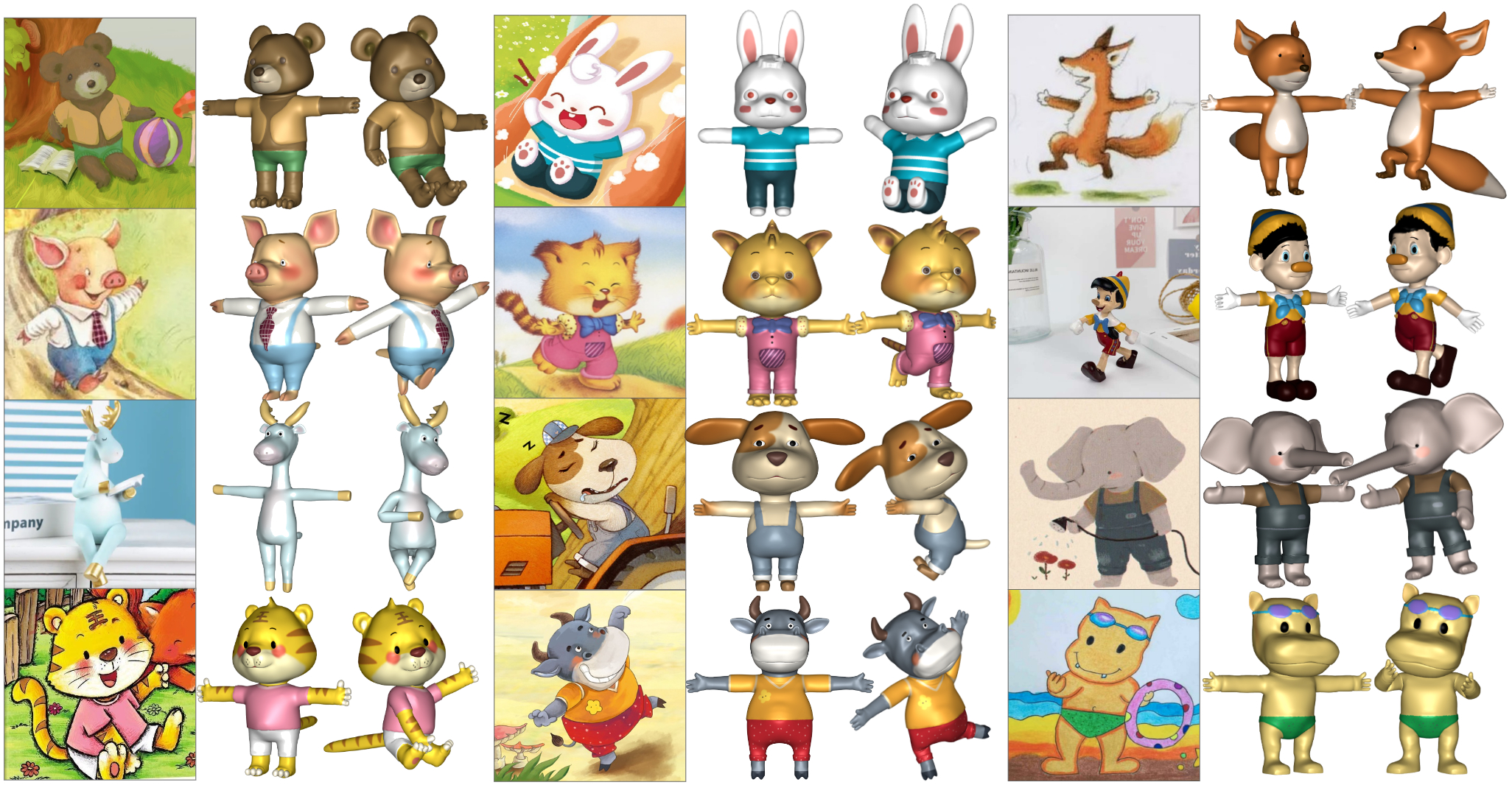}
  \caption{The \textbf{gallery} of the representative examples sampled from \textit{3DBiCar}. Each collected reference image is followed by the T-pose model and the posed model, created by professional artists. \textit{3DBiCar} contains 1,500 topologically consistent, textured and skinned 3D high-quality models with paired 2D images, which covers 15 species and 4 image styles.
  %Dataset Gallery of \textit{3DBiCar}: Here are 12 examples from all styles of images in \textit{3DBiCar}. Each example is represented by three graphs, a 2D picture, a 3D T-Pose model, and a 3D posed model, respectively, from left to right.
  }
  \label{fig_dataset_gallery}
\end{figure*}

%% file: figure/fig_table.tex
\begin{figure}[htb]
  \centering
  \includegraphics[width=.98 \linewidth]{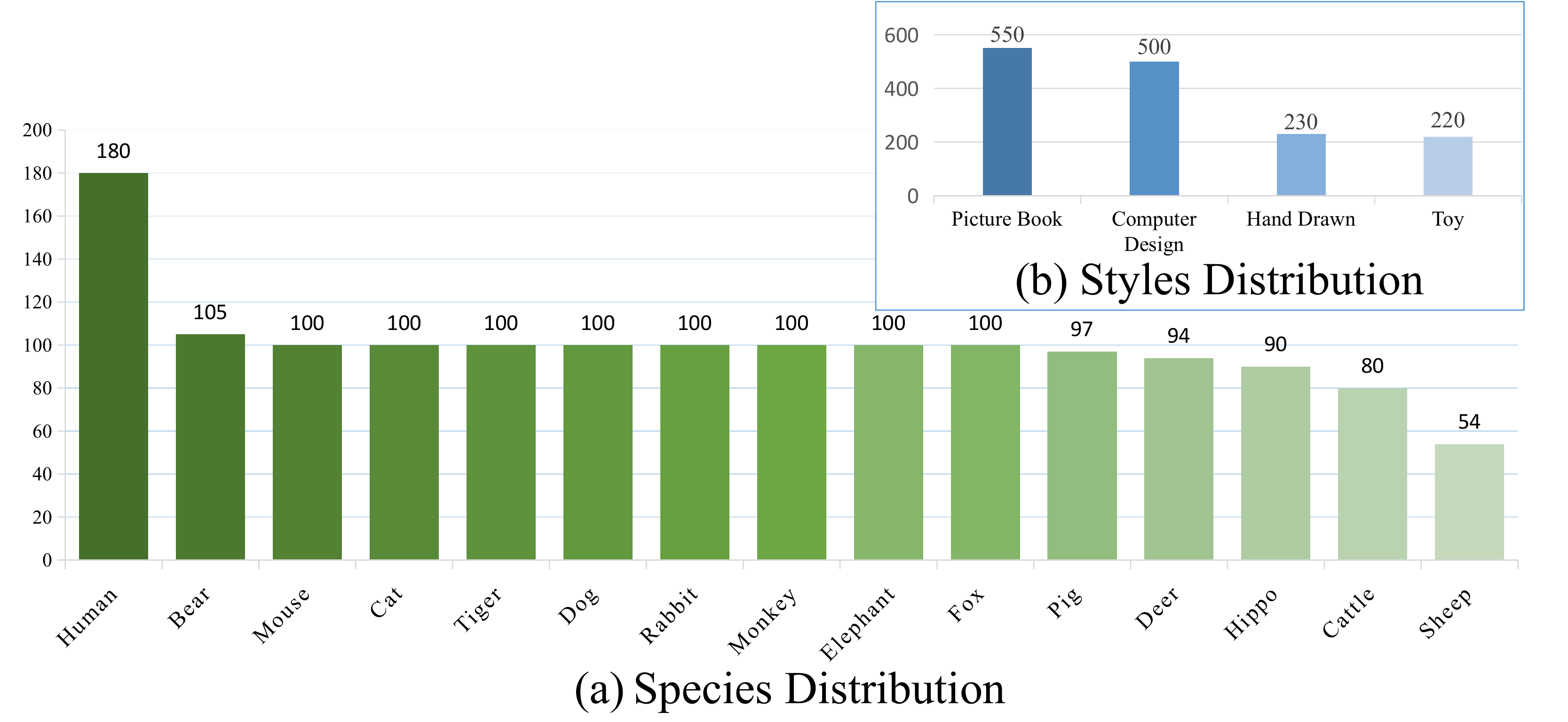}
  \caption{\textbf{Data distribution.} Chart (a) illustrates the number of 15 species of bipedal cartoon characters in \textit{3DBiCar}. Chart (b) shows the number of four styles of reference images collected in our dataset.}
  \label{fig_datainfo}
\end{figure}

%% file: figure/fig_dataset_pipeline.tex
% \iffalse
% \begin{figure*}[htb]
%   \centering
%   \includegraphics[width= .98 \linewidth]{image/fig_dataset_pipeline.pdf}
%   \caption{Dataset Building Pipeline: The generation process of a biped character model from an image.
%   }
%   \label{fig_dataset_pipeline}
% \end{figure*} \fi

\begin{figure}[htb]
  \centering
  \includegraphics[width=.98 \linewidth]{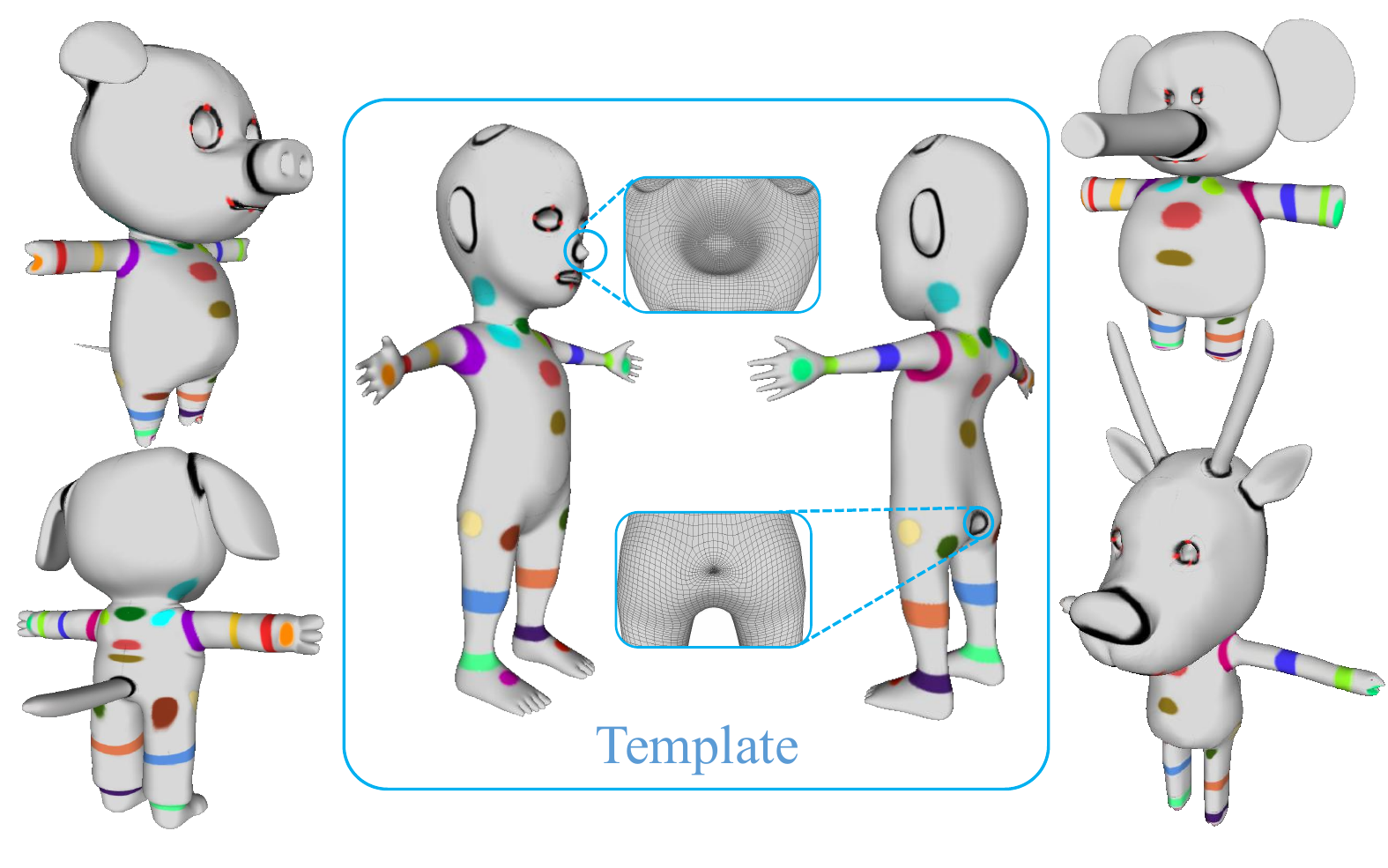}
  \caption{\textbf{Template.} The models in the center are the predefined template mesh with landmarks. It can be seen that we refine the structure on specific regions, where a complex nose or tail may exist. The colored regions and delineated lines denote the landmarks. These landmarks represent specific components of the character's body, such as elbow and eye socket. During model crafting, artists are required to deform the template model while keeping the landmarks in the position where the original body components are.
  }
  \vspace{-0.3cm}
  \label{fig_template_design}
\end{figure}

%% file: paper/04_method.tex
\section{Parametric Modeling}
% \section{\textit{BiCar}}
\label{sec:algorithm}
% \input{figure/fig_SMCL_process}
%It is a challenging task for a neural network to fit our cartoon models with texture due to numerous diverse data (up to $10^7$ parameters), even though the models are topologically consistent.

We propose the first pa\textbf{ra}metric model of 3D \textbf{bi}ped car\textbf{t}oon characters (\textit{RaBit}), which contains a linear blend model for shapes and a neural generator for textures. \textit{RaBit} simultaneously parameterizes the shape, pose, and texture of 3D biped characters. Specifically, we decompose the parametric space into identity-related body parameter $B$ (Sec.~\ref{shape}), non-rigid pose-related parameter $\Theta$ (Sec.~\ref{pose}) and texture-related parameter $T$ (Sec.~\ref{texture}). Overall, a 3D biped character is parameterized as follows,
\begin{equation}
\begin{split}
    M =& F(B, \Theta, T) \\
      =& F_T(F_P(F_S(B),\Theta), T),
\label{eq:1}
\end{split}
\end{equation}
where $F_S$, $F_P$, and $F_T$ are the parametric functions to generate shape, pose, and texture respectively. The following sections will elaborate on the details of \textit{RaBit}.

% where $F_S$, $F_P$, and $F_T$ are the parametric functions to generate shape, pose, and texture respectively, as illustrated in Fig.~\ref{fig_SMBL_process}. The following sections will elaborate the details in our parameterization.

\subsection{Shape Modeling}
\label{shape}
Recently, linear shape models dominate the representation of statistical 3D model. Numerous methods~\cite{SMPL-X:2019, blanz1999morphable, cao2013facewarehouse, FLAME:SiggraphAsia2017} have shown PCA's ability in modeling the human body and face. Inspired by~\cite{SMPL:2015}, we parameterize our character shape linearly with the following equation, 
\begin{equation}
    M_{S} = F_S(B) = \bar{M}_{S}+\sum_i^{|B|} \beta_i s_i,
\end{equation}
where $\bar{M}_{S}$ denotes the mean shape and ${M}_{S}$ is the reconstructed shape. The coefficients of linear shape are $\beta_i \in B$. $|B|$ is the number of shape parameters and is set to 100 in our implementation. $s_i \in \mathbb R^{3 \times N}$ denotes the orthogonal principal components of vertex displacements that capture shape variations in different character identities. The shape model of \textit{RaBit} is learned from 1,050 characters of \textit{3DBiCar} using PCA~\cite{SMPL:2015}. RaBit's eyeballs can be computed based on the predefined landmarks shown in Fig.~\ref{fig_template_design}. Please refer to the Supplementary for more details.

\subsection{Pose Modeling}
\label{pose}
\textit{RaBit} employs a standard vertex-based linear blend skinning technique, which uses the predefined skeleton and skinning weight matrix provided by \textit{3DBiCar}. The pose parameter $\Theta$ defines a set of angles as $\Theta=[\theta_1,\theta_2,...,\theta_K]\in \mathbb{R}^{69}$, where $\theta_k \in \mathbb{R}^{3}$ denotes the axis-angle representation of the relative rotation of joint $k$ with respect to its parent and $K=23$ represents the number of joints. $\theta_k$ can be converted to the rotation matrix format $R(\theta_k)$ using Rodrigues' formula~\cite{SMPL:2015}.
%The pose parameter $\Theta$ defines a set of angles as $\Theta=[\theta_1,\theta_2,...,\theta_K]\in \mathbb{R}^{69}$, where $K = 23$ represents the number of joints. The rotation of node $k$ can be expressed as $\theta_k \in \mathbb{R}^{3}$ where $\theta_k$ can be converted to rotation matrix format $R(\theta_k)$ using Rodrigues' formula. 
The following equations demonstrate how the pose function $F_P$ changes vertex $v_i \in M_{S}$ to its corresponding position ${v_i}^\prime \in M_{P}$,
\begin{equation}
    {v_i}^\prime = \sum^K_{k=1} w_{k, i} G_k'(\Theta, J) v_i,
\end{equation}
\begin{equation}
    G_k'(\Theta, J) = G_k(\Theta, J) G_k(\Theta', J)^{-1},
\end{equation}
\begin{equation}
    G_k(\Theta, J) = \prod_{j\in A(k)} 
    \left[ \begin{array}{cc}
        R(\theta_j)  & J_j \\
        0 & 1
    \end{array}\right],
\end{equation}
where $w_{k,i}$ is the $k$-th element of the linear blend matrix $W$ for the $i$-th vertex. $G_k(\Theta, J)$ is the global transformation of joint $k$, while $G_k'(\Theta, J)$ is the global transformation of joint $k$ after removing the transformation of rest pose $\Theta'$. $A(k)$ denotes a set including all the ancestors of joint $k$, and $J_j$ represents the location of the $j$-th joint, which is located at the bounding box center of a specific body landmark. Thus given the T-pose mesh $M_S$ and specific pose $\Theta$, we can generate the corresponding posed mesh $M_P$ by $F_P$,
\begin{equation}
    M_P = F_P(M_S,\Theta).
\end{equation}

\subsection{Texture Modeling}
Although traditional linear PCA is capable of building a decent statistical shape model, it fails to represent high-frequency details in textures and can produce blurry results due to its weak Gaussian assumption. Recently, GAN-based architectures~\cite{wang2018pix2pixHD,karras2020analyzing,karras2019style,fu2022stylegan,grigorev2021stylepeople,gecer2019ganfit} have shown the notable capability of generating high-fidelity images. Thus, we resort to StyleGAN2-based techniques for UV texture maps generation, but with a coherent UV unfolding (as shown in Fig.~\ref{fig_pipeline_svr}) to facilitate the learning of texture compared with~\cite{grigorev2021stylepeople}. Specifically, the neural texture generator in \textit{RaBit} translates a latent code to a texture map, which could be formulated as follows,
\begin{equation}
    G\left(T\right): \mathbb{R}^{Z} \rightarrow \mathbb{R}^{H \times W \times C},
\end{equation}
where $G\left(T\right)$ takes a Z-dimensional parameter vector as input and synthesizes a $H \times W\times C$ texture map. Thus given a posed mesh $M_P$ and a specific texture code $T$, we can generate a textured mesh $M$ with the following equation,
\begin{equation}
    M = F_T(M_P,T) = H(M_P, G\left(T\right)),
\end{equation}
where $H$ means the process of applying the texture map to the mesh model. In our implementation, the generator follows the architecture of StyleGAN2~\cite{karras2020analyzing},  while taking a 512-dimensional parameter vector as input and generating a $1024\times1024\times3$ texture map.

\label{texture}

%% file: paper/05_application.tex
\section{Single-View Reconstruction}
\label{sec:experiment-svr}
\input{figure/fig_pipeline_svr}

Single-view reconstruction (SVR) is one of the most popular tasks of efficient 3D content generation, and recent work has made noticeable progress on human reconstruction based on parametric model of human characters (e.g., SMPL). To verify the practicality of our proposed \emph{3DBiCar} and \emph{RaBit}, we conduct SVR for bipled cartoon characters. A baseline learning-based method is presented, which is termed as \textit{BiCarNet}. 

\subsection{\textit{\textbf{BiCarNet}}}

Given a single masked image of cartoon characters, our \textit{BiCarNet} aims to reconstruct the corresponding 3D shape, pose, and texture. %As \textit{Rabit} spans a large space of textured models. 
The key problem is to build an encoder to map the input image to the parametric space of \textit{Rabit}. As shown in the upper part of Fig.~\ref{fig_pipeline_svr}, we adopt the learning block in HMR~\cite{kanazawa2018end} as our Encoder. Once these parametric vectors are learned, we can feed them to our \textit{RaBit} model to generate a posed body mesh, two eyeballs, and a UV texture (we name it global for convenience to introduce the following method description). 

During our preliminary experiments, we find that the shape reconstruction of characters, \ie the eyes and body, is satisfactory, while the inferred UV tends to lose detailed appearances of some small yet significant areas, such as the nose and ears. We thus adopt a part-sensitive texture reasoner (PSR) to address the above issue, as the lower part of Fig.~\ref{fig_pipeline_svr} shows. Specifically, we design five individual UV-mappings for these significant parts of the nose, ears, horns, eyes, and mouth. Five lightweight encoder-decoder branches are next introduced to learn the appearances of these local regions from the input image, respectively. The learned part UVs could be remapped to the corresponding area on the global UV map to produce a blended texture. However, a direct blending tends to cause seam artifacts. Thus we further adopt a Fuser to address the artifacts as Fig.~\ref{fig_pipeline_svr} illustrates. Please refer to the Supplementary for thorough implementations of \textit{BiCarNet}.

\subsection{Experiments}

\textbf{Data preparation.} We first split \textit{3DBiCar} into a training set (1,050 image-model pairs) and a testing set (450 pairs). To support a stable training of \textit{BiCarNet}, we next generate a large number of synthetic paired data with the help of \textit{RaBit}, which are highly diversified in shape, posture, and texture. To be specific, a lot of 3D textured models are first sampled from the \textit{RaBit} space, which are then rendered into images from different camera views. 
This finally produce 13,650 pairs for training. Note that, \emph{BiCarNet} takes an image with foreground masked as input. All synthetic images naturally have no background. For real images, all the foreground masks are manually annotated. 

\textbf{Result gallery.} 
Fig.~\ref{fig_wildresult} shows representative results generated by \textit{BiCarNet}. As illustrated, our \textit{BiCarNet} can generate vivid 3D cartoon characters loyal to individual cartoon images in shape, pose, and texture. We believe that our work opens the door to producing 3D biped cartoon characters from easy-to-obtain inputs.
\input{figure/wild_result}

\textbf{Results on Shape Reconstruction.} 
As mentioned above, \textit{BiCarNet} utilizes HMR-like blocks and \textit{RaBit} for shape and pose learning. Currently, other reconstruction methods could also be used for topology-consistent geometry inference, such as GCNN-based methods~\cite{lin2021-mesh-graphormer} and UV-based methods~\cite{zeng20203d}. We choose two representative methods for comparison, i.e., Mesh-Graphormer~\cite{lin2021-mesh-graphormer,lin2021end-to-end} and DecoMR~\cite{zeng20203d}. Mesh-Graphormer combines graph convolutions and self-attentions in a transformer for 3D human reconstruction from a single image. DecoMR reconstructs 3D human mesh from single images by regressing a UV-based location map. Tab.~\ref{tab_mesh_result} shows the quantitative comparisons of the above three methods on MPVE, MPJPE, and PA-MPJPE. We also provide qualitative comparisons in Fig.~\ref{fig_mesh_result}. Both quantitative and qualitative results demonstrate that the HMR-like method achieves the highest performance on geometry inference and provides more accurate reconstructions closer to ground truths. As noted, both Mesh-Graphormer and DecoMR outperform HMR for SMPL-based human reconstruction. It is interestingly found that they perform worse in our settings. One possible reason is that our biped cartoon meshes own an extremely larger amount of vertices than SMPL to model more complex geometry. This greatly increases the challenge of vertices regression in Mesh-Graphormer and DecoMR. Thus, in our setting, directly regressing the low-dimension parameters performs better.

\input{table/table_mesh_result}
\input{figure/fig_mesh_result}

\textbf{Results on Texture Inference.} 
To demonstrate the capability of our proposed GAN-based texture generator, we first compare our \emph{RaBit}-based texture inference (i.e., \emph{BiCarNet}) with PCA-based inference. Specifically, for PCA-based method, we utilize the same learning architecture to map the input image into the PCA-based texture space. Furthermore, %with the traditional PCA-based method. Specifically, 
to evaluate the effectiveness of the proposed texture inference modules, we also conduct ablative analysis on \textit{BiCarNet} without Fuser and \textit{BiCarNet} without Part-sensitive Reasoner (PSR). Table~\ref{tab_texutre_result} shows the quantitative results of different texture inference methods on MSE, PSNR and FID and our proposed method achieves the highest scores compared with all other methods. Moreover, Fig.~\ref{fig_texture_result} illustrates the qualitative results of these methods for texture inference. Fig.~\ref{fig_texture_fusion} shows the results without and with Fuser, which demonstrates our fusion module can deal with unnature seam-like artifacts. We can observe that the part-sensitive texture reasoner and the Fuser help to capture the local regions of characters and recover their detailed appearances. 

\input{table/table_texture_result}
\input{figure/fig_texture_result}
\input{figure/fig_texture_fusion.tex}

\section{More Applications}
\label{sec:application}
\subsection{Sketch-based Modeling}
Customizing 3D biped cartoon characters usually requires a heavy workload with commercial tools, even for experienced artists. Sketch-based modeling enables amateur users to get involved in 3D shape customization in a simple and intuitive fashion. We build a sketch-based modeling application with the help of \textit{3DBiCar} and \textit{RaBit}. 

We first sample a series of shape vectors randomly and feed them to \textit{RaBit} to generate 3D cartoon characters with diversified shapes, resulting in 12,000 T-pose models. Then we apply suggestive contour~\cite{han2017deepsketch2face} to render the front-view sketches with different abstraction levels and obtain 108,000 sketch-model pairs. Given a sketch as input, we employ ResNet-50 and three MLPs as the encoder and decoder to map the input sketch to 100-dimensional shape parameters. The generated shape parameters are next fed to \textit{RaBit} to reconstruct the corresponding 3D model. Please refer to the Supplemental materials for more details. Note that users only need to depict a 3D character with T-pose on a 2D canvas while the output characters of our method are animation-ready and can be directly applied to other commercial tools. Fig.~\ref{fig_sketch} displays the sketches created by users with little knowledge of modeling as well as the corresponding models generated by our system. It can be seen that our sketch-based modeling system offers a smart approach for amateur users to create 3D biped cartoon characters with diversified shapes. We will further explore the support of shape reconstruction from  sketches with arbitrary poses, and texture painting in the future.

\input{figure/sketch_result}

\subsection{3D Character Animation}
Following the recent advance of human recovering method and parametric model~\cite{wang2022live,SMPL:2015,SMPL-X:2019}, we extract the human from input video frames and adopt a temporal-aware encoder to recover the sequence of human poses~\cite{wang2022live}. Then, a motion retargeting method~\cite{hsieh2005motion} is used to convert the poses on the human skeleton to the motion of our cartoon characters. As shown in Fig.~\ref{fig_anim}, animation-ready characters generated by our \textit{RaBit} can be directly applied to 3D animation. Please refer to the supplementary for animation video.

\input{figure/fig_anim}

%% file: figure/fig_pipeline_svr.tex
% \begin{figure*}[htbp]
%   \centering
%   \includegraphics[width=0.92 \linewidth]{image/fig_pipeline2.pdf}
%   \caption{%The architecture of our baseline method. 
%   \textbf{Single view reconstruction baseline.}
%   Given an image as input, we firstly adopt two CNNs and three MLPs as the encoders, mapping the input to three low-dimension latent codes respectively, i.e., the parameters of shape, pose and texture. Then these three parameters are fed into our \textit{SMCL} model to generate the predicted mesh with texture and pose. Finally, we take the ground truth model from \textit{3DBiCar} as supervision to train our neural network. Our \textit{SMCL} is not only responsible for the shape, pose and texture generation, but also takes charge of the eyes computing.
%   %Overview of our baseline pipeline: Input images are fed to the encoder, and output shape, pose, and texture parameters. With \textit{SMBL}, posed textured models are reconstructed with predicted parameters.} %In our pipeline, we invlove in prevent shape and pose parameter from influencing mutually
%   }
%   \label{reconstruction:pipeline}
% \end{figure*}

\begin{figure}[htbp]
  \centering
  \includegraphics[width=.96\linewidth]{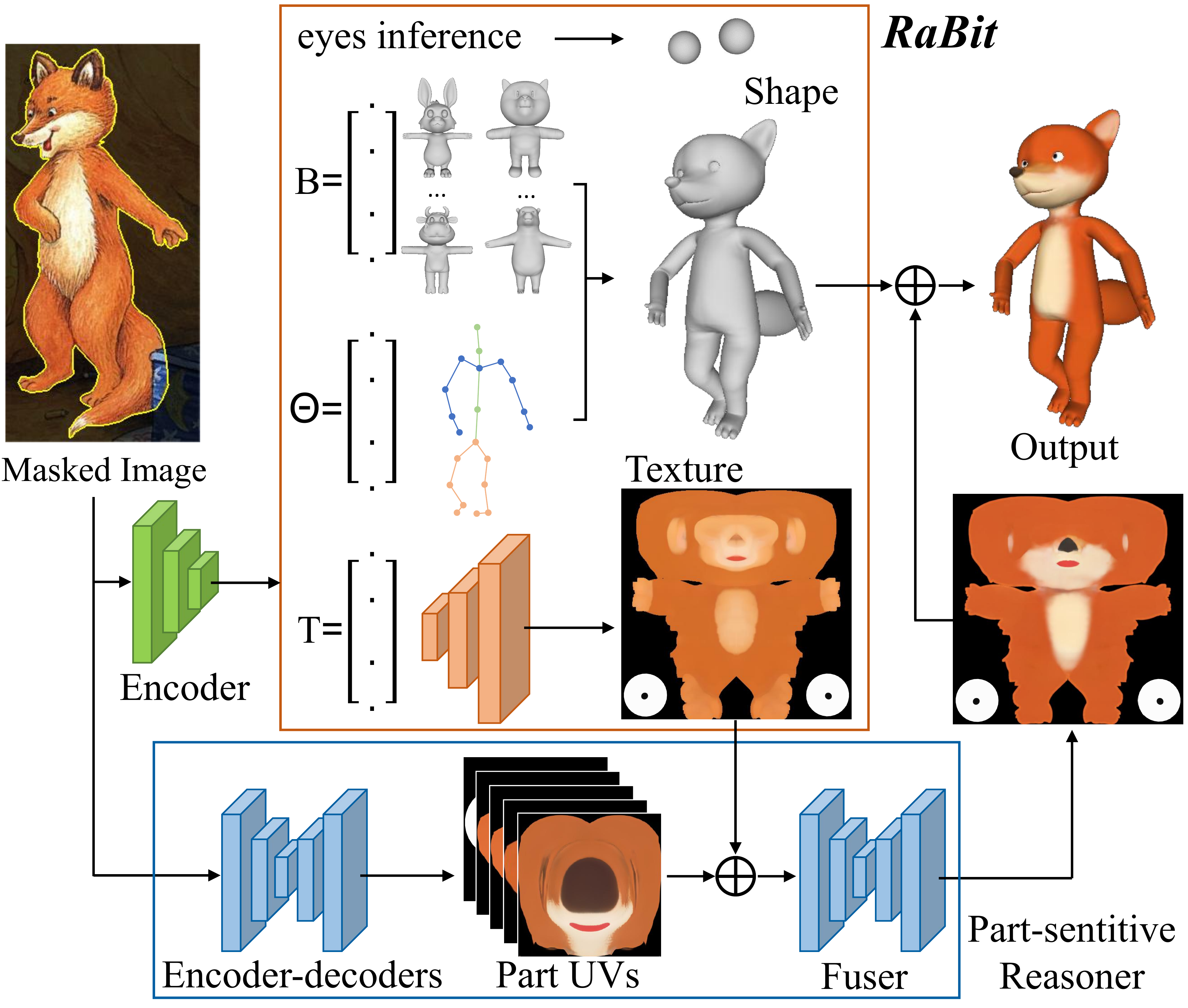}
  \caption{\textit{\textbf{BiCarNet}.} Given a masked image, we first map it to the parametric vectors. The vectors are then fed to our \textit{RaBit} to generate a posed body mesh, two eyeballs, and a global UV texture. A part-sensitive reasoner is utilized to perceive local regions and generate the detailed UV texture map. Finally, a vivid 3D cartoon character is obtained with our \textit{BiCarNet}.}
  \label{fig_pipeline_svr}
\end{figure}

%% file: figure/wild_result.tex
\begin{figure}[htbp]
  \centering
  \includegraphics[width=.98 \linewidth]{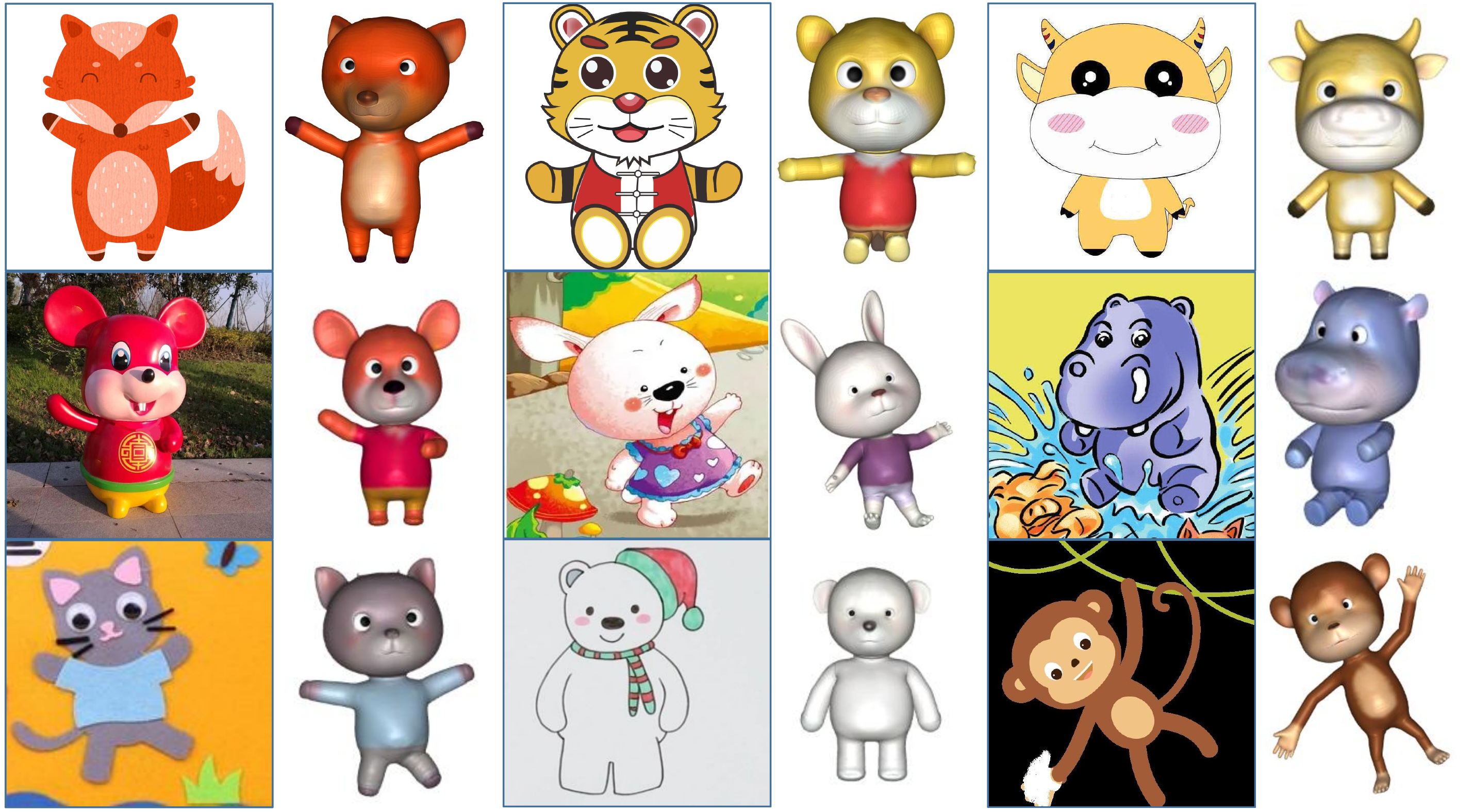}
  \caption{\textbf{Result gallery of \textit{BiCarNet}.} Our \textit{BiCarNet} is capable of generating vivid 3D cartoon characters with only single-view image input.
  }
  \label{fig_wildresult}
\end{figure}

%% file: table/table_mesh_result.tex
%\iffalse
\begin{table}[htbp]
    \renewcommand{\arraystretch}{1.3}
    \centering
    \resizebox{\columnwidth}{!}{
        \begin{tabular}{l|c|c|c}
        % \hline 
        \toprule
        Method & MPVE $\downarrow$ & MPJPE $\downarrow$ & PA-MPJPE $\downarrow$ \\
        \midrule
        DecoMR\cite{zeng20203d} & 85.74 & 81.23 & 47.23\\
        Mesh-Graphormer\cite{lin2021-mesh-graphormer} & 63.31 & 47.15 &	34.12\\
        Ours (HMR\cite{kanazawa2018end} + \textit{RaBit}) & \textbf{51.46} & 	\textbf{37.80} & \textbf{25.97}\\
        \bottomrule
        \end{tabular}
    }
    \caption{\textbf{Quantitative results of shape reconstruction.} Our method achieves the best results in terms of MPVE, MPJPE and PA-MPJPE. Note that all metrics are measured in a unit $10^{-3}$m.} %P-MPJPE reports the MPJPE error after alignment with 3d keypoint of ground truth in translation and rotation. Theta reports MAE between predition and ground truth. T-rec reports the point-wise MSE between predition and ground truth model with Tpose. Projection error between manual 2D annotation on image and 2D projections of models after alignment
    \label{tab_mesh_result}
\end{table}
%\fi

% 	input	mPVE	mPJPE	PAmPJPE
% hmr	224	9.28164	7.73912	--
% Mesh Graphormer	224	63.31	47.15	34.12
% decomr	224	19.0676	6.967964	4.84606

%% file: figure/fig_mesh_result.tex
\begin{figure}[htp]
  \vspace{-0.2cm}
  \centering
  \includegraphics[width= 0.98\linewidth]{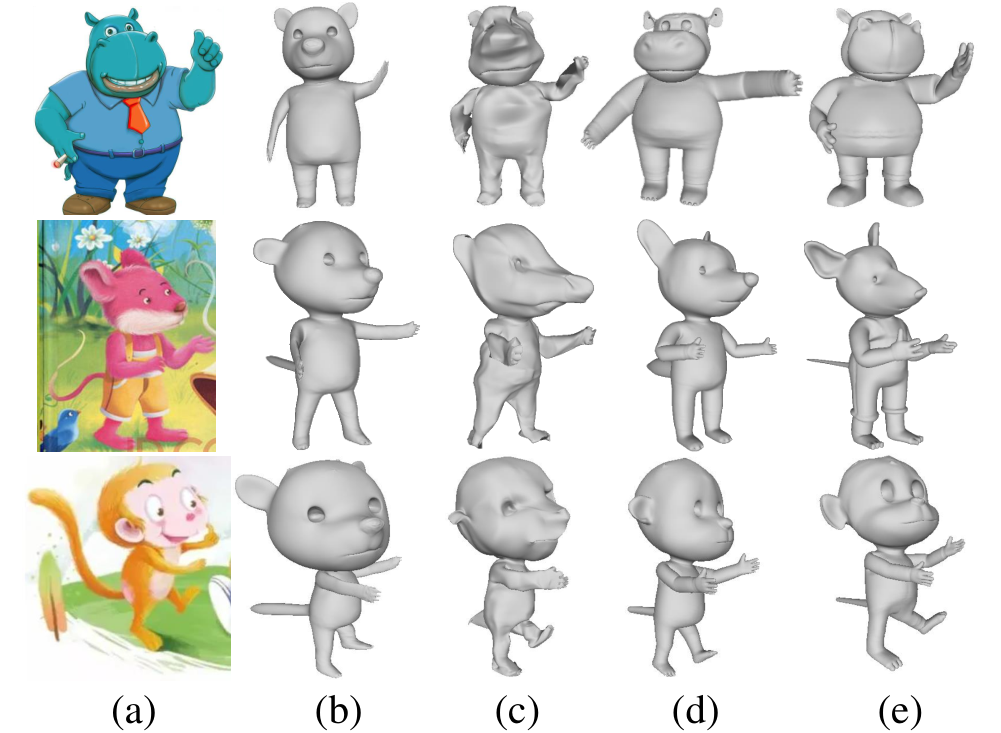}
  \caption{\textbf{Qualitative results of shape reconstruction.} From left to right, each row contains (a) the input image, reconstructed meshes of (b) Mesh Graphormer, (c) DecoMR, (d) our method, and (e) the GT mesh.}
  \vspace{-1.2mm}
  \label{fig_mesh_result}
\end{figure}

%% file: table/table_texture_result.tex
%\iffalse
\begin{table}[htbp]
    \renewcommand{\arraystretch}{1.3}
    \centering
    \resizebox{0.45\textwidth}{!}{
        \begin{tabular}{l|c|c|c}
        % \hline 
        \toprule
        Method & MSE($\times 10^{-1}$)$\downarrow$ & PSNR($\times 10^2$) $\uparrow$ & FID $\downarrow$ \\
        \midrule
        PCA  & 0.2309 & 0.2254 & 0.4642 \\
        % \hline
        % StyleGan  & 
        % \textbf{0.1630} & \textbf{0.2578} & \textbf{0.3553} \\
        % \midrule
        \textit{BiCarNet}  & \textbf{0.1093} & \textbf{0.2458} & \textbf{0.1133} \\
        
        \midrule
        \textit{BiCarNet} w/o Fuser  & 0.1108 & 0.2397 & 0.1407 \\
        \textit{BiCarNet} w/o PSR  & 0.1346 & 0.2361 & 0.4024 \\
        % \hline
        % BP  & 0.1328 & 0.2364 & 0.3103 \\
        % \hline
        % BPH  & 0.1325 & 0.2349 & 0.2716 \\
        % \hline
        % pSp  & 0.1281 & 0.2285 & 0.4869 \\
        % \midrule
        
        % \hline
        % p2p_2  & 0.1373 & 0.2272 & 0.1107 \\
        % \hline
        \bottomrule
        \end{tabular}
        %result is check!
    }
    \caption{\textbf{Quantitative results on texture inference.} PCA denotes linear modeling method for texture and the last two rows indicate the results of \textit{BiCarNet} respectively without two designed module. Our \textit{BiCarNet} outperforms others methods in all metrics.}
    \label{tab_texutre_result}
\end{table}
% \vspace{-0.6cm}
%\fi

%% file: figure/fig_texture_result.tex
\begin{figure}[htp]
  \vspace{-0.2cm}
  \centering
  \includegraphics[width=1.0 \linewidth]{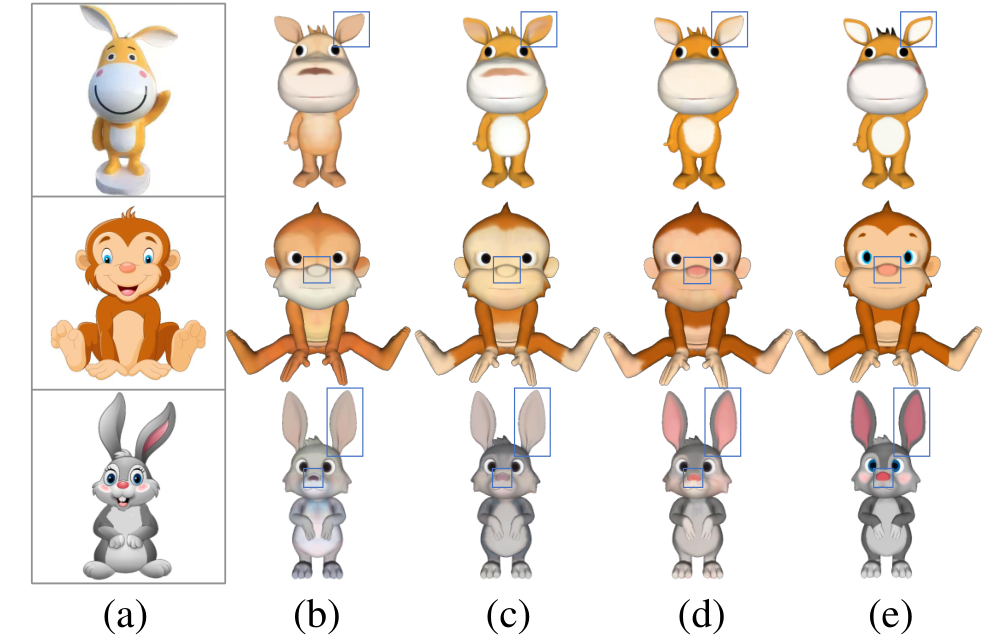}
  \caption{\textbf{Qualitative comparisons on texture inference.} The input image (a) is followed by the textured models from (b) PCA, (c) \textit{BiCarNet} w/o PSR, (d) \textit{BiCarNet} and (e) the ground truth. Note that we use the same shape and focus on the difference of textures.}
  \label{fig_texture_result}
\end{figure}

%% file: figure/fig_texture_fusion.tex
\begin{figure}[htbp]
%   \vspace{-0.8cm}
  \centering
  \includegraphics[width=.95\linewidth]{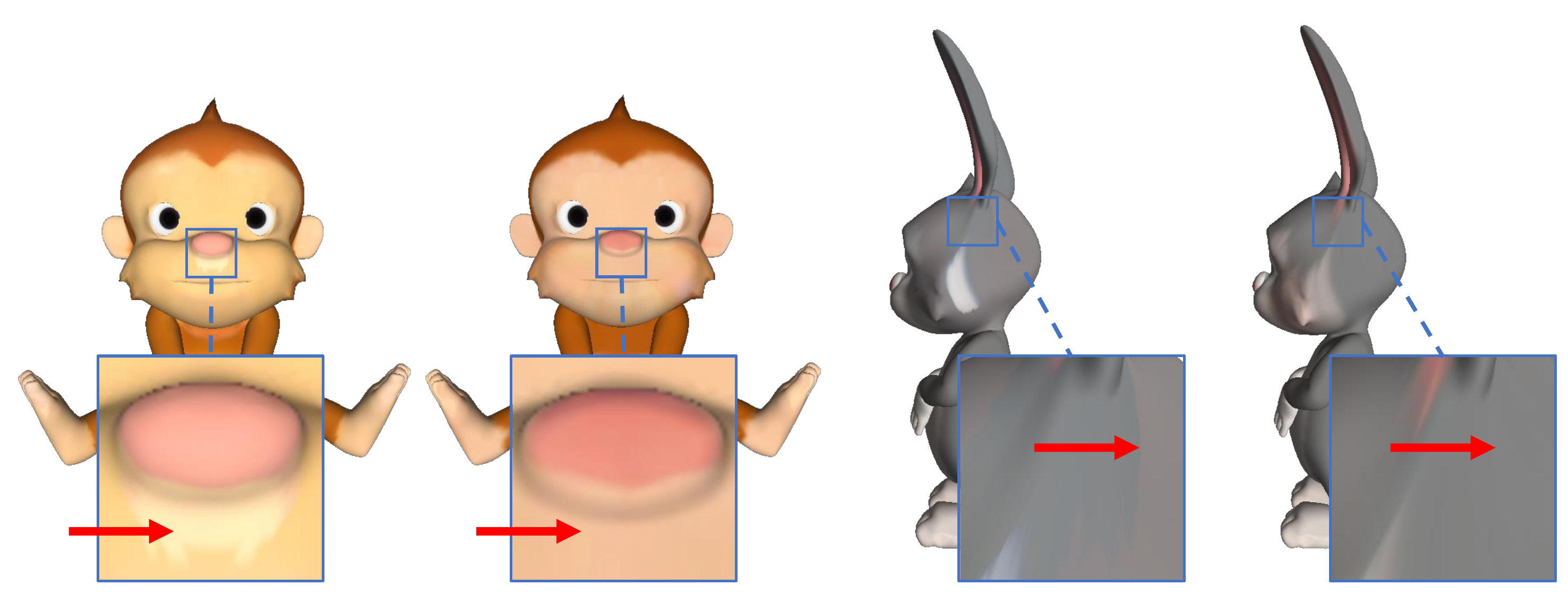}
  \caption{\textbf{Qualitative ablation on Fuser in Texture inference.} Left: \textit{BiCarNet} w/o Fuser. Right: \textit{BiCarNet} with Fuser.}
  \label{fig_texture_fusion}
\end{figure}

%% file: figure/sketch_result.tex
\begin{figure}[ht]
  \centering
  \includegraphics[width=.99\linewidth]{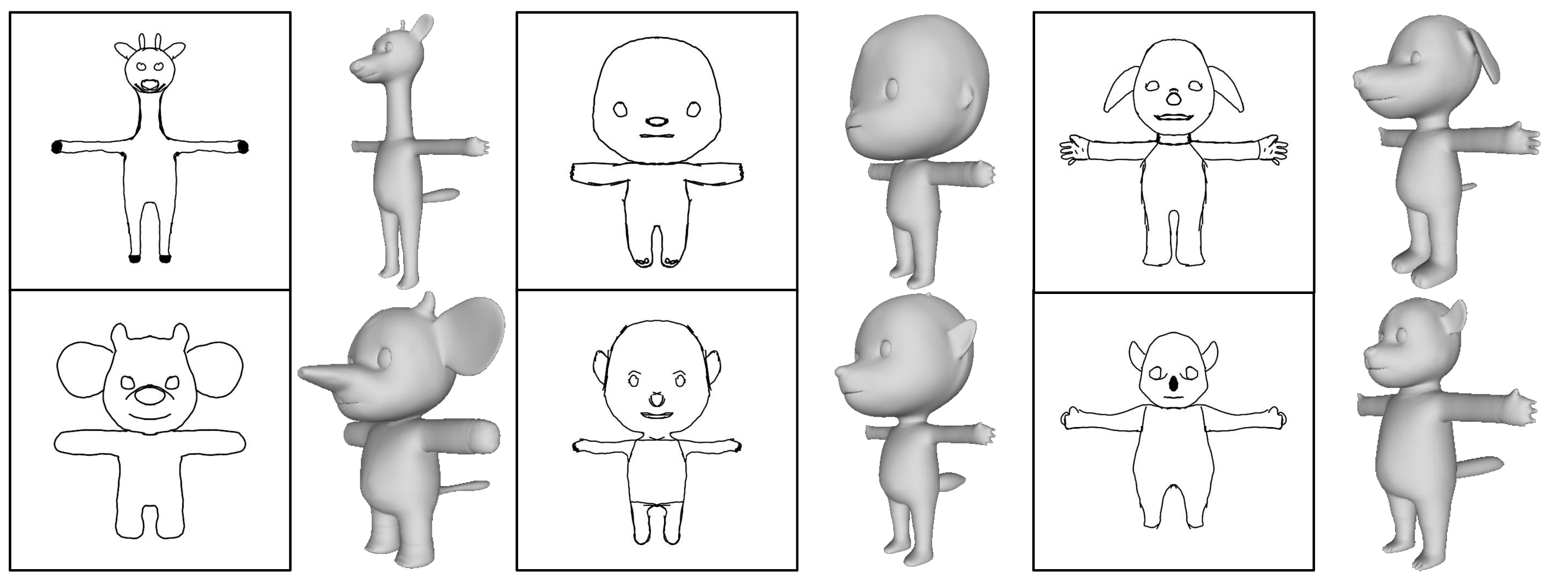}
  \caption{\textbf{Result gallery of sketch-based modelling.} The sketches created by amateur users denotes on the left and the generated models on the right.}
  \label{fig_sketch}
\end{figure}
\vspace{-0.5cm}

%% file: figure/fig_anim.tex
\begin{figure}[htbp]
  \centering
  \includegraphics[width=.98 \linewidth]{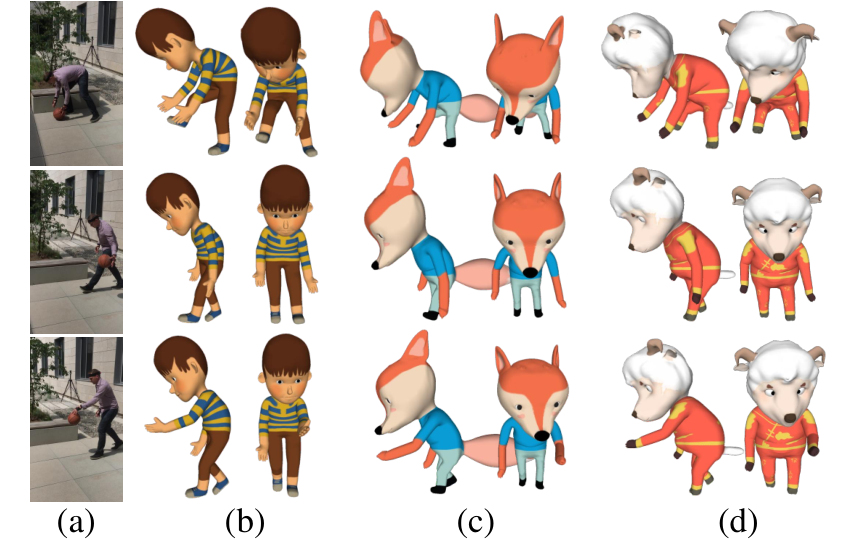}
  \caption{\textbf{Transferring motion of a human video to animate characters.} (a) denotes the input frames. (b), (c), and (d) indicate three corresponding posed cartoon characters.}
  \label{fig_anim}
\end{figure}
\vspace{-0.3cm}

%% file: paper/07_conclusion.tex
\section{Conclusion}
\label{sec:conclusion}

In this work, we introduce \textit{3DBiCar}, the first large-scale 3D biped cartoon character dataset. It contains 1,500 textured and skinned models with a consistent mesh topology. Based on \textit{3DBiCar}, we propose the first 3D full-body cartoon parametric model \textit{RaBit} for biped character modeling. Furthermore, we build a baseline method \textit{BiCarNet} for reconstructing 3D textured models from a single image with cartoon characters. Experimental results demonstrate the capability of \textit{3DBiCar} and \textit{RaBit} as well as the effectiveness of \textit{BiCarNet}. Last but not least, two further applications, i.e., sketch-based modeling and 3D character animation, demonstrate the usability and practicality of our dataset and parametric model. %We hope that our work can open a door for further researches in the area of cartoon character digitalization.
We hope that our work will contribute to the development of 3D biped cartoon character modeling and inspire future works in this area.

\noindent\textbf{Acknowledgements.} The work was supported in part by NSFC with Grant No.~62293482, the Basic Research Project No.~HZQB-KCZYZ-2021067 of Hetao Shenzhen-HK S\&T Cooperation Zone. It was also partially supported by Shenzhen General Project with No.~JCYJ20220530143604010, the National Key R\&D Program of China with grant No.~2018YFB1800800, by Shenzhen Outstanding Talents Training Fund 202002, by Guangdong Research Projects No.~2017ZT07X152 and No.~2019CX01X104, by the Guangdong Provincial Key Laboratory of Future Networks of Intelligence (Grant No.~2022B1212010001), and by Shenzhen Key Laboratory of Big Data and Artificial Intelligence (Grant No.~ZDSYS201707251409055).

%% file: 12_appendix.tex
% \input{_constants}
% %\review % \review OR \cameraready
% \cameraready
% % \documentclass[10pt,twocolumn,letterpaper]{article}

% % \newcommand{\zj}[1]{{\color{blue}            {#1}}}

% % \input{cvpr_header}
% % \myexternaldocument{_main}
% \appendix
% %%

% % \begin{document}
% \input{supp/3DBiCar}
% \input{supp/RaBit}
% \input{supp/BiCarNet}
% \input{supp/Sketch}
% % \end{document}

% %optional
% % \input{supp_figure/fig_interpolation}
% % \input{supp_figure/fig_assemble}
% % \input{supp_figure/fig_poseaug}
% % \input{supp_figure/fig_uv_interpolation}
% % \input{supp/sample_smcl}

\appendix
\label{sec:appendix}
\input{supp/3DBiCar}
\input{supp/RaBit}
\input{supp/BiCarNet}
\input{supp/Sketch}

%% file: supp/3DBiCar.tex
\section{Details of \textit{3DBiCar}}
\noindent\textbf{Image Styles.} Fig.~\ref{image_style} shows representative images for the 4 styles with different appearances. We define them based on their different sources: \textit{picture book} - cropped from e-books of children, \textit{computer designed} - made by artists using software, \textit{hand drawn} - drawn by kids, \textit{toy} - captured from real toys.

\input{rebuttal_figure/fig_style}

\noindent\textbf{Shape Modeling Procedure.} We recruit six professional artists to create 3D corresponding character models using Blender according to the collected reference images. The key to building a linear parametric shape model lies in maintaining a unified mesh topology. To achieve this, all six artists are required to craft 3D models by deforming the template mesh under the constraints of the predefined landmarks. Each artist owns over six years of modeling experience and each character takes around 1 hour on average. The modeling result is required to be matched with the reference images as much as possible. To maintain visual quality and topological consistency, we have established a review committee of ten members to assess the models based on reference images and predefined landmarks.

%% file: rebuttal_figure/fig_style.tex
\begin{figure}[htb]
  \centering
  \includegraphics[width=.98\linewidth]{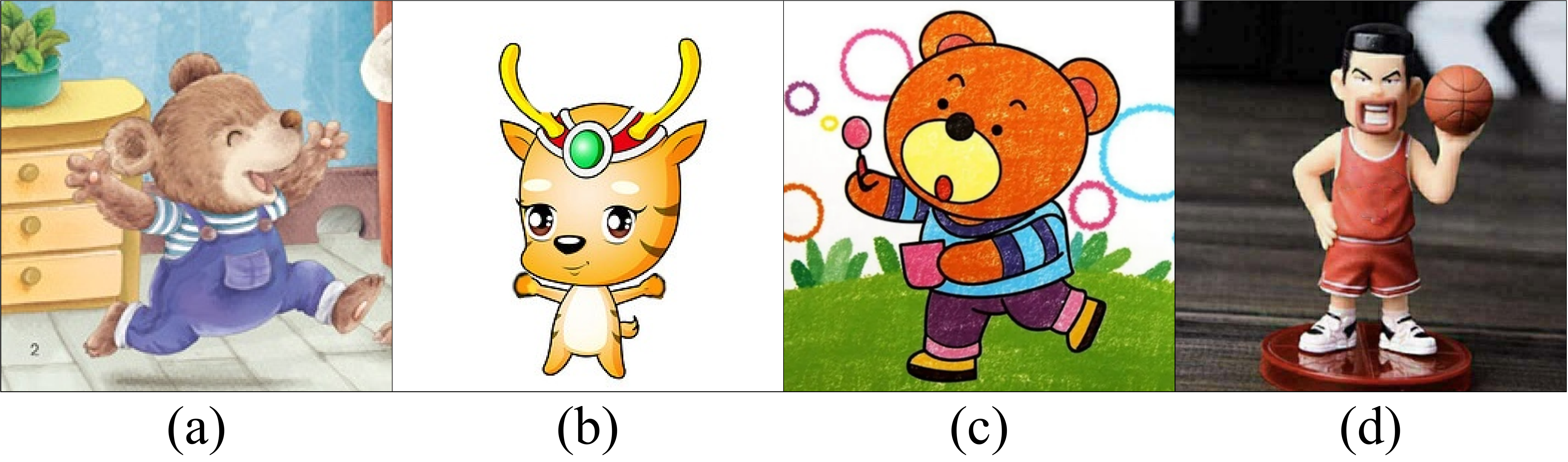}
  \caption{A representative example from different image styles. (a) picture book, (b) computer designed, (c) hand drawn, (d) toy.}
  \label{image_style}
\end{figure}

%% file: supp/RaBit.tex
\section{Details of \textit{RaBit}}

\noindent\textbf{Shape Space.} As illustrated in Fig.~\ref{rec}, \textit{RaBit} is able to express the \textit{basic geometry} of diverse shapes in \textit{3DBiCar} with \textit{low-dimensional vectors} (100 in our experiments). Such ability of \textit{RaBit} can well facilitate the construction of learning-based regression methods for inferring reasonable shapes from images or sketches, as demonstrated in our downstream tasks. Biped cartoon is known as a popular character style in gaming and filming. \textit{RaBit} spans a wider range of species than existing human model~\cite{SMPL:2015,SMPL-X:2019}. However, due the use of the holistic PCA model, \textit{RaBit} may struggle to represent local geometric details and may result in undesirable entanglement, as shown in Fig.~\ref{axis}. Conducting parametric modeling for diverse shapes is a fundamental problem, but it has received little methodological evolution in the past due to the lack of data. We hope that our proposed dataset can inspire further research in this area.

% We agree that the specific type limits the usability (new types require new designs, i.e., quadruped), while this in contrast provides feasibility for parametric modeling. Even so, the problem is still very challenging. Biped cartoon is known as a popular character style in gaming and filming. Our model spans a wider range of species than existing human model. We hopes to take a small step in this field.  
%our experiments show that L-PCA is able to express the \textit{basic geometry} of diverse shapes with \textit{low-dimensional vectors} (100 in our experiments), which is verified by Fig.~\ref{rec}. %The first two shape principal components of shape are shown in .
\input{rebuttal_figure/fig_rec}

\input{rebuttal_figure/fig_axis}

\noindent\textbf{Visualization of Topological Consistency.} Good correspondence of training data is essential for constructing a linear shape model and preserving the topological consistency of reconstructed models. Getting topological consistency in manual modeling is intrinsically challenging. To do so, we put much effort to construct \textit{3DBiCar}, including template designing, landmark guidance, review committee for careful checking. In Fig.~\ref{correspondence}, we use checkboard texture mapping for visualizing the correspondence of representative examples sampled from \textit{RaBit}'s shape space.

\input{rebuttal_figure/fig_correspondance}

\noindent\textbf{Eyeball Reconstruction.}
In our implementation, we approximate an eyeball as a sphere. Generally, a sphere is determined by its center and radius. As shown in the Fig.~\ref{eyeball}, in \textit{RaBit}, an eyeball's center $\mathbf{o_{e}}$ and radius $r_{e}$ is computed as follows,

\begin{equation}
    r_{e} = c_1 r_s,
\end{equation}
\begin{equation}
    d_{e} = c_2 r_s,
\end{equation}
\begin{equation}
    \mathbf{o_e} = \mathbf{o_s} - d_{e} \mathbf{n},
\end{equation}
where $r_s$ and $\mathbf{o_s}$ is the radius and the center of the 3D circle, computed by the least square fitting with the landmark points of the eye socket. $d_e$ denotes the Euclidean Distance between $\mathbf{o_s}$ and $\mathbf{o_e}$ and $\mathbf{n}$ the normal of the 3D circle. $c_1$ is the mean value of ${d_e}/{r_s}$ of all models in \textit{3DBiCar},  while $c_2$ is the mean value of ${r_e}/{r_s}$. Both $c_1$ and $c_2$ are precomputed constant values.

\input{supp_figure/eye_ball}

\noindent\textbf{Implementations.} The shape model of \textit{RaBit} is learned from 1,050 models of \textit{3DBiCar} using PCA~\cite{pedregosa2011scikit,SMPL:2015}.
For pose modeling, \textit{RaBit} utilizes the consistent skeleton and skinning weight matrix defined in \textit{3DBiCar}. Note that both \textit{3DBiCar} and \textit{RaBit} currently does not support the animation of tails, which will be explored in our future work. As for texture modeling, 1,050 raw textures from \textit{3DBiCar} were adopted and extended to 21,000 training data with image-level augmentations (e.g., flipping, and adjusting HSV). \textit{Rabit}'s texture generator follows the architecture of StyleGAN2~\cite{karras2020analyzing} and is trained with the following setting:  the dimensionality of $Z$ with $512$, the output resolution with $1024 \times 1024 \times 3$, the learning rate with $3 \times 10^{-4}$, the batch size with 32, the Adam optimizer with $\beta_1=0$, $\beta_2=0.99$, $\epsilon=10^{-8}$. The training is performed on a server with 4 Nvidia RTX 3090Ti GPUs.

%% file: rebuttal_figure/fig_rec.tex
\begin{figure}[htb]
  \centering
  \includegraphics[width=1.\linewidth]{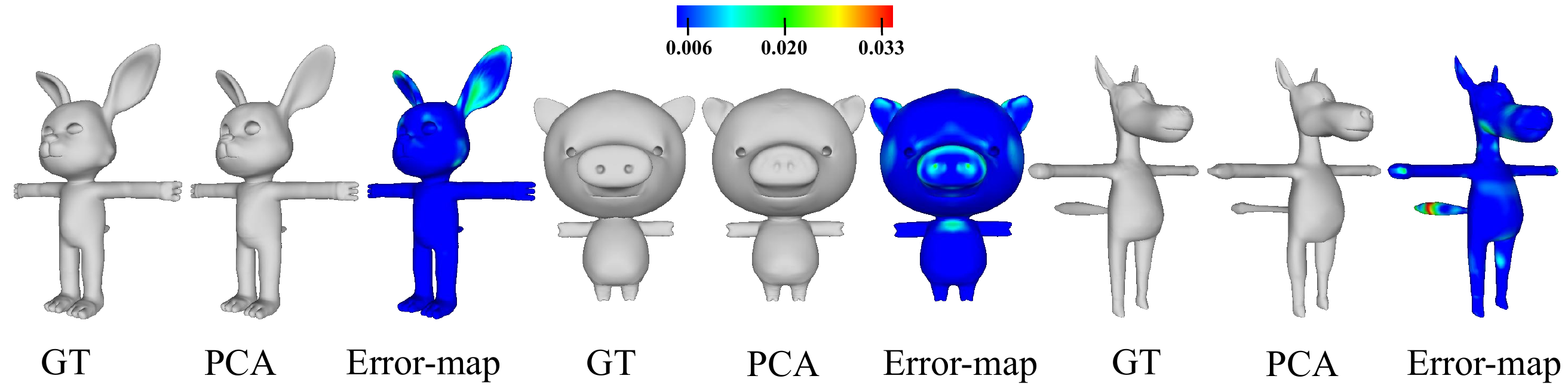}
  \caption{Comparison of shapes reconstructed by \textit{RaBit} with GT.}
  \label{rec}
\end{figure}

%% file: rebuttal_figure/fig_axis.tex
\begin{figure}[htb]
  \centering
  \includegraphics[width=0.98\linewidth]{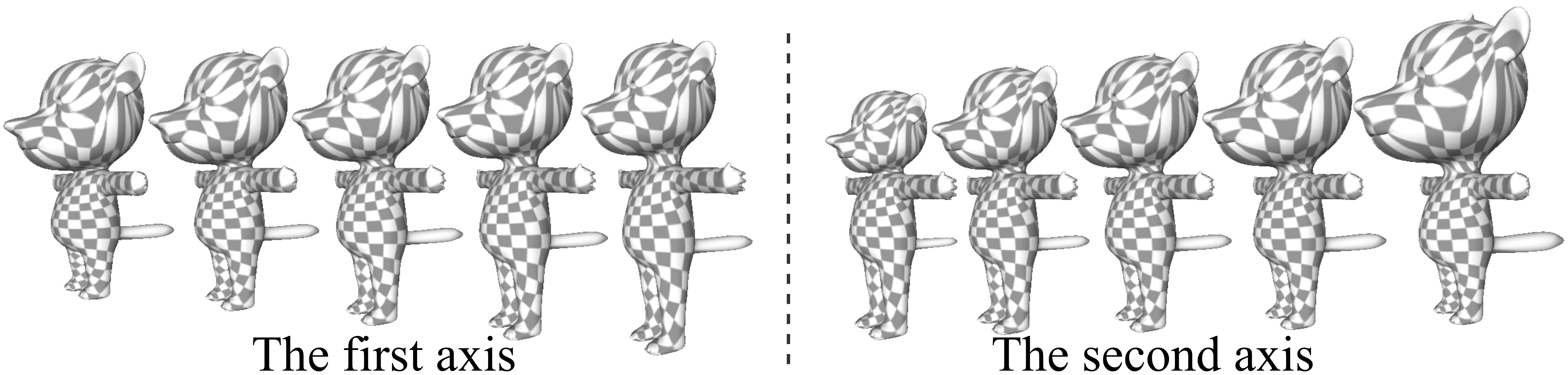}
  \caption{An illustration of the first two axes of shape space in \textit{RaBit}.}
  \label{axis}
\end{figure}

%% file: rebuttal_figure/fig_correspondance.tex
\begin{figure}[htb]
  \centering
  \includegraphics[width=.98\linewidth]{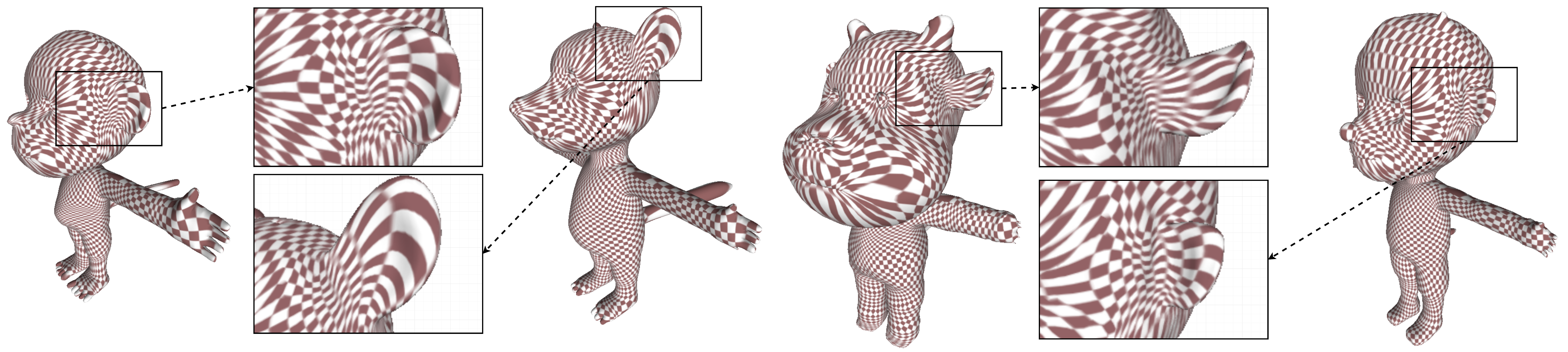}
  \caption{An illustration of the mesh correspondence.}
  \label{correspondence}
\end{figure}

%% file: supp_figure/eye_ball.tex
\begin{figure}[htb]
  \centering
  \includegraphics[width=.68 \linewidth]{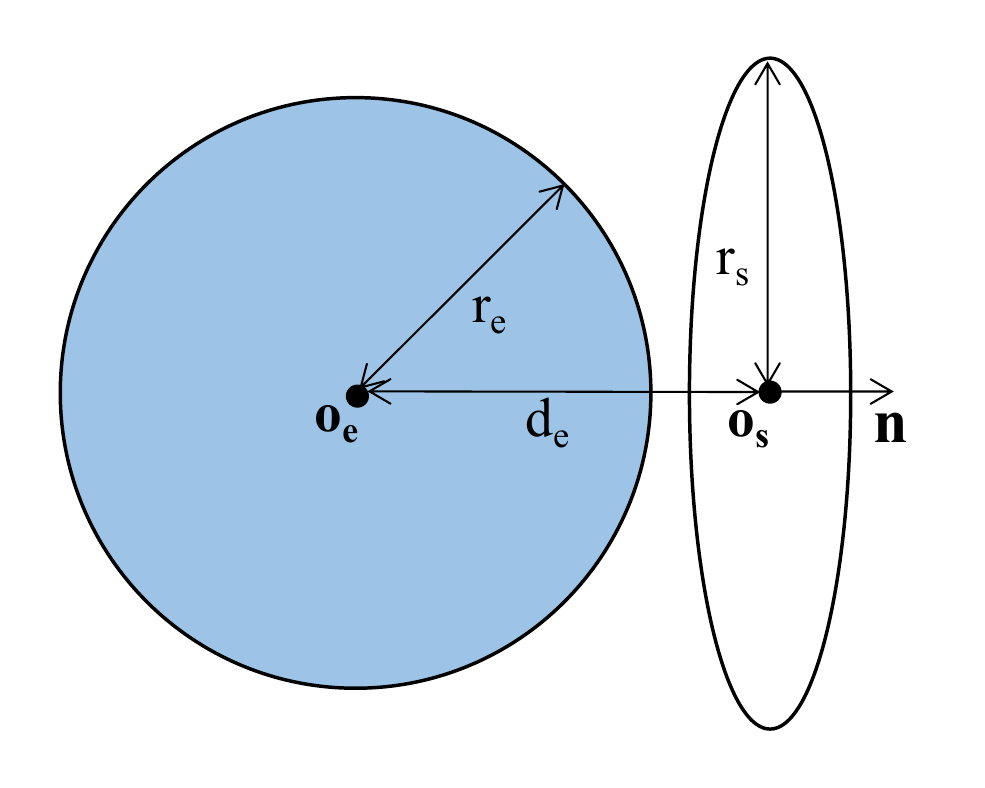}
  \caption{An illustration of eyes computation. $\mathbf{o_e}$ is the center of the eye and $r_e$ is the radius of the eye. $\mathbf{o_s}$ and $r_{s}$ are the center and the radius of the orbit, respectively. }
  \label{eyeball}
\end{figure}

%% file: supp/BiCarNet.tex
\section{Details of \textit{BiCarNet}}

\noindent\textbf{Data Preparation.} 
We split \textit{3DBiCar} into a training set (1,050 image-model pairs) and a testing set (450 pairs). To support a stable training of \textit{BiCarNet}, we augment a large number of synthetic paired data with the help of \textit{RaBit}. Specifically, we generate a series of shape vectors by interpolating between the 1,050 models' shape parameters. Fig.~\ref{fig_interpolation} shows the representative results of interpolated shapes. For pose augmentation,  a variety of poses from other datasets (e.g., Human3.6M\cite{Ionescu2014Human36M}) are retargeted to \textit{RaBit}'s pose space, as shown in Fig.~\ref{fig_poseaug}. Furthermore, 1,050 raw textures are also utilized to generate synthetic texture maps by interpolating with \textit{RaBit}, as shown in Fig.~\ref{fig_uv_interpolation}. The above augmentations finally produce 13,650 models with texture and pose. These models are then rendered into images from different camera views for training. 

\input{supp_figure/fig_interpolation}
\input{supp_figure/fig_poseaug}
\input{supp_figure/fig_uv_interpolation}

\noindent\textbf{Implementations.} In our implementation, for the shape and pose regression modules, we utilize two ResNet-50 blocks to embed the input image ($512 \times 512 \times 3$) to a 100-dimensional shape vector and a 69-dimensional pose vector, respectively. For the texture module, we adopt pSp-encoder~\cite{richardson2021encoding} to learn a 512-dimensional texture vector from the image. As for the part-sensitive texture reasoner, we use pSp~\cite{richardson2021encoding} as the basic building block and learn multiple local UV textures ($256 \times 256 \times 3$) from the input. pix2pixHD~\cite{wang2018pix2pixHD} is employed as the fusion module (Fuser), which takes the $1024 \times 1024 \times 3$ coarsely-blended texture map as input and outputs fine texture maps with the same resolution.

\noindent\textbf{Part-Sensitive UVs.} As shown in Fig.~\ref{uv_style}, we design five individual UV-mappings for significant parts, i.e., nose, ears, horns, eyes, and mouth. These part UVs enlarge five constant regions of the global UV mapping. Five lightweight encoder-decoder branches are adopted to learn the appearances of these local regions from the input image, respectively. The learned part UVs could then be remapped to their corresponding areas on the global UV map, resulting in a blended texture.

% More detailed descriptions will be included in the revision. Here, we illustrate the part UV layouts and textures in the left part of Fig.~\ref{uv_style} to give a brief explanation. These part UVs enlarge five constant regions of the global UV mapping. 

\input{rebuttal_figure/fig_uv}

%% file: supp_figure/fig_interpolation.tex
\begin{figure}[htbp]
  \centering
  \includegraphics[width=.96\linewidth]{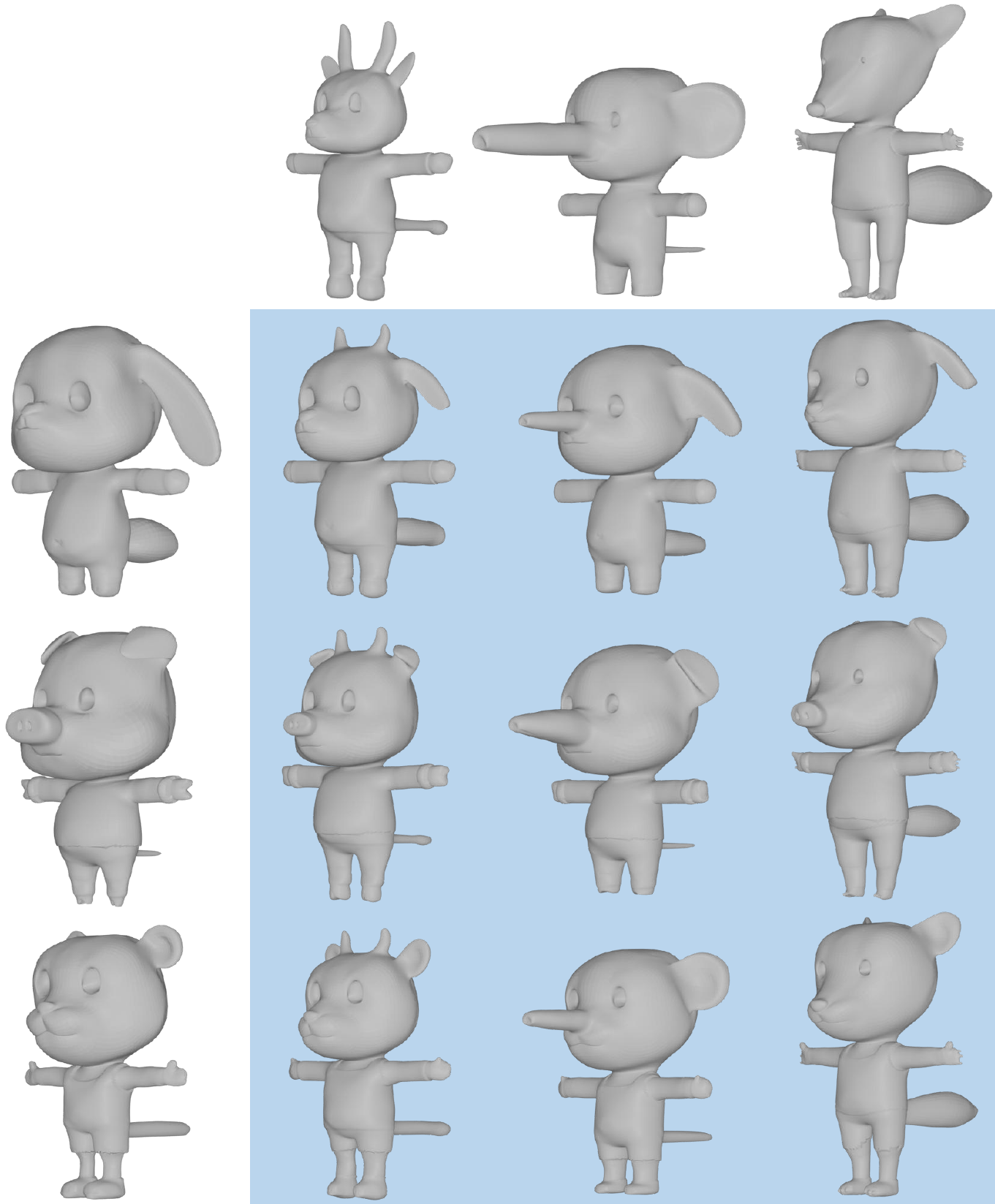}
  \caption{An illustration of interpolated shapes. Models from the top row and left column are from \textit{3DBiCar}. Other models with blue backgrounds are obtained by interpolating the leftmost and uppermost models with the help of \textit{RaBit}.}
  \label{fig_interpolation}
\end{figure}

%% file: supp_figure/fig_poseaug.tex
% \begin{figure}[htbp]
%   \centering
%   \includegraphics[width=.98 \linewidth]{placeholder/fig_poseaug.pdf} %pose aug
%   \caption{Pose Sampling: Tpose indicates T-pose model. Pose\* indicate pose of coresponding model in \textit{3DBiCar}. Pose1 and Pose2 is two poses transfer from Human3.6M \cite{Ionescu2014Human36M}}
%   \label{fig_poseaug}
% \end{figure}

\begin{figure}[htbp]
  \centering
  \includegraphics[width=.90 \linewidth]{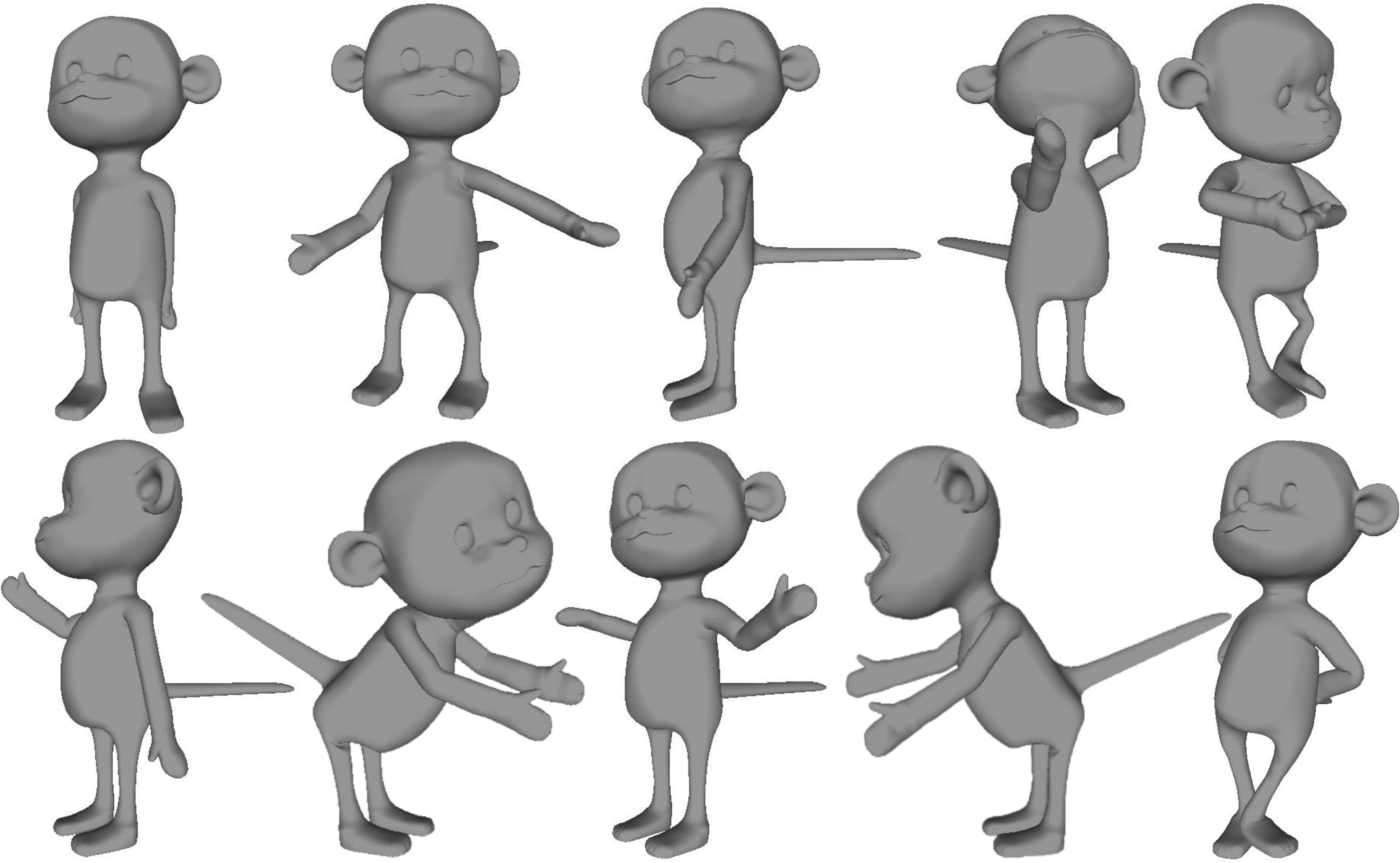} %pose aug
%   \caption{Diverse poses from  Human3.6M \cite{Ionescu2014Human36M} are transfered to a model in \textit{3DBiCar}.}
  \caption{An illustration of diverse poses transferred from pose datasets.}
  \label{fig_poseaug}
\end{figure}

%% file: supp_figure/fig_uv_interpolation.tex
\begin{figure}[htbp]
  \centering
  \includegraphics[width=.98 \linewidth]{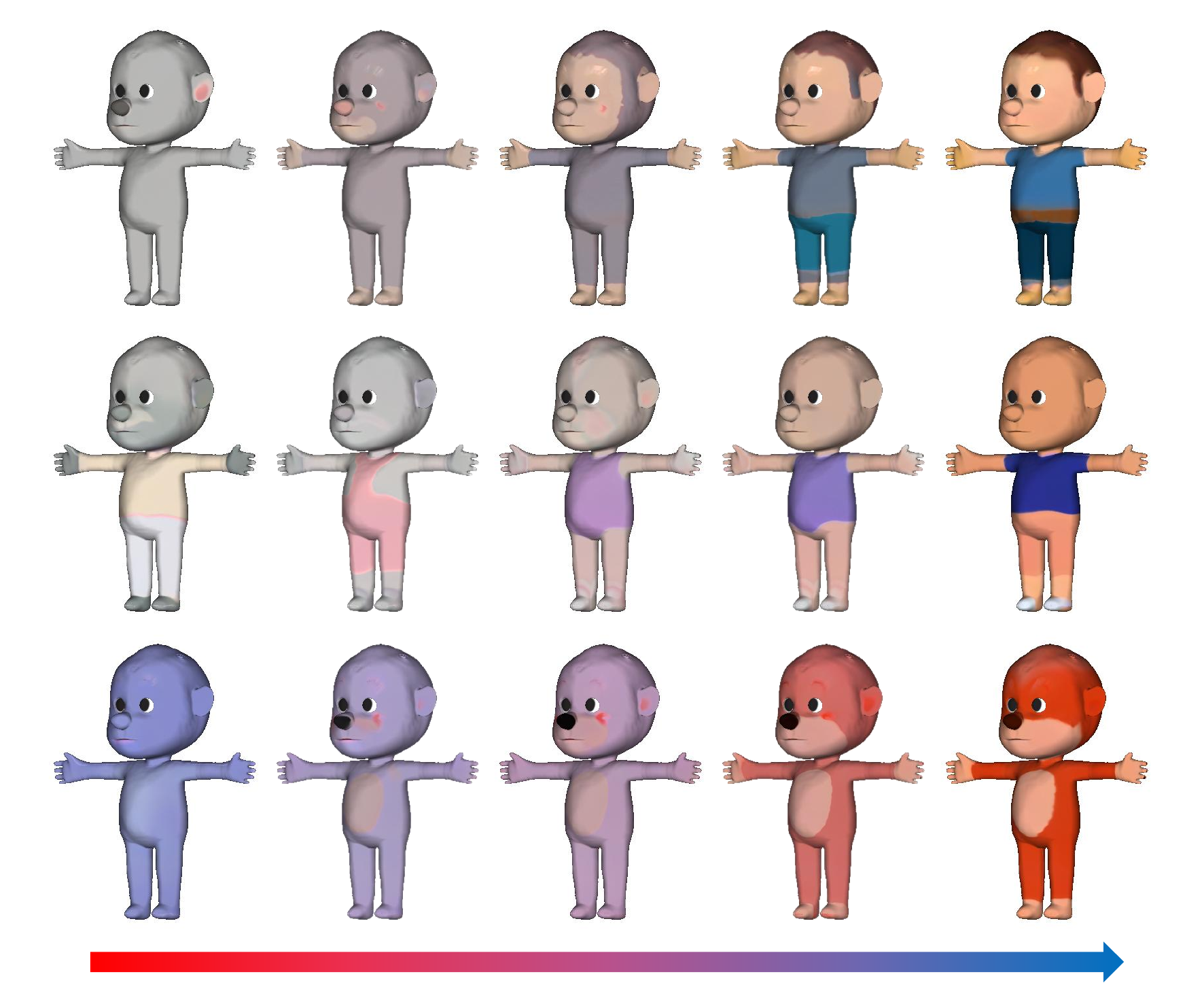} %assemble.png
  \caption{An illustration of synthesized texture maps. For each row, the leftmost and the rightmost textures are from \textit{3DBiCar}, while the other three textures are interpolated results generated by \textit{RaBit} under different weights.}
  \label{fig_uv_interpolation}
\end{figure}

%% file: rebuttal_figure/fig_uv.tex
\begin{figure}[htb]
  \centering
  \includegraphics[width=0.96\linewidth]{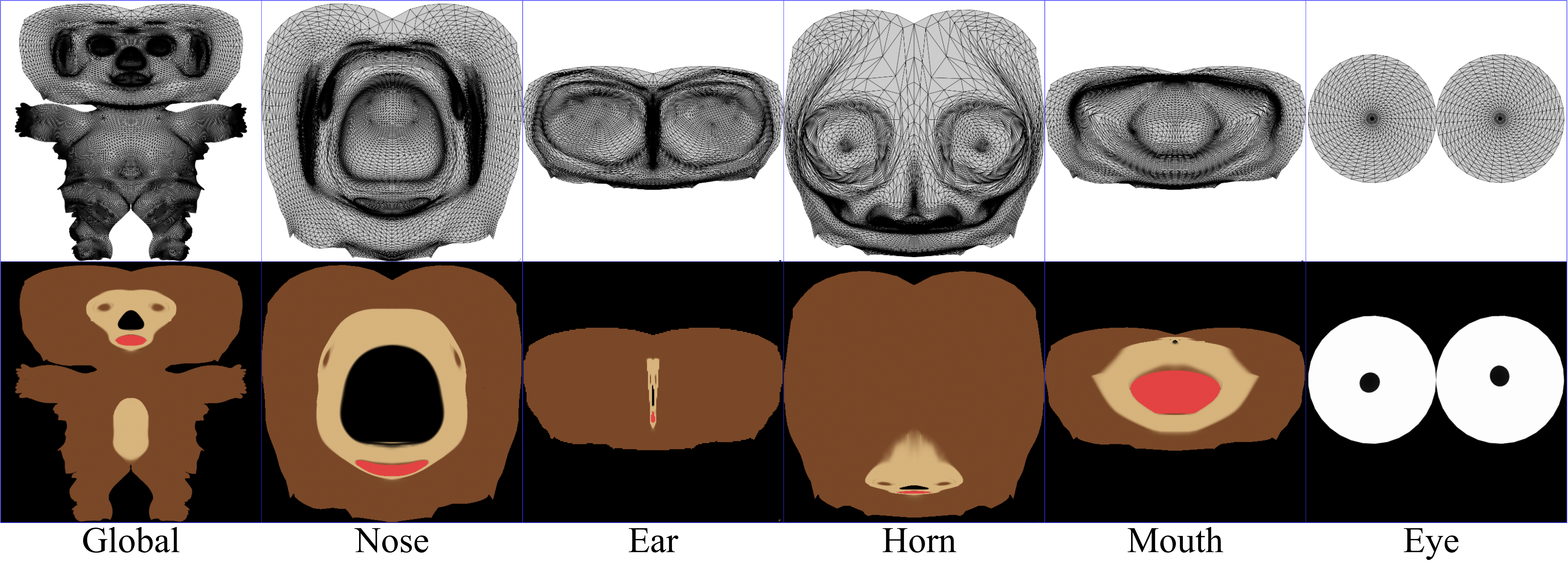}
  \caption{An illustration of our UV layouts and textures.}
  \label{uv_style}
\end{figure}

%% file: supp/Sketch.tex
\section{Details of Sketch-based Modeling}

\noindent\textbf{Data preparation.} We first sample 12,000 shape vectors randomly and feed them to \textit{RaBit} to generate 3D cartoon characters with diversified shapes. Then the suggestive contour~\cite{decarlo2003suggestive, han2017deepsketch2face} is applied to render the front-view sketches with different abstraction levels and obtain 108,000 sketch-model pairs. Fig.~\ref{fig_sketch_data} shows examples of rendered sketches. 

\input{supp_figure/fig_sketch_data}

\noindent\textbf{Implementations.} As shown in Fig.~\ref{fig_sketch_pipeline}, we first adopt one ResNet-50 module and three MLPs as the encoder-decoder architecture, mapping the input sketch $512 \times 512$ to 100-dimensional shape parameters. Then the generated shape parameters are fed to \textit{RaBit} to reconstruct the corresponding 3D model. We train the network with a batch size of $100$ and a learning rate of $3 \times 10^{-4}$ with the Adam optimizer. Moreover, we use the $L_1$ loss to measure the difference between the predicted shape parameters and the ground truth. Our sketch-based modeling interface is implemented with the QT framework. CGAL is adopted for 3D geometry processing. As shown in the video, running on a personal computer with an Intel i7-7700 CPU, 16GB RAM, and a single Nvidia GTX 2080Ti GPU, our modeling application supports real-time feedback. 

\input{supp_figure/fig_sketch_pipeline}

%% file: supp_figure/fig_sketch_data.tex
\begin{figure}[htbp]
  \centering
  \includegraphics[width=.98\linewidth]{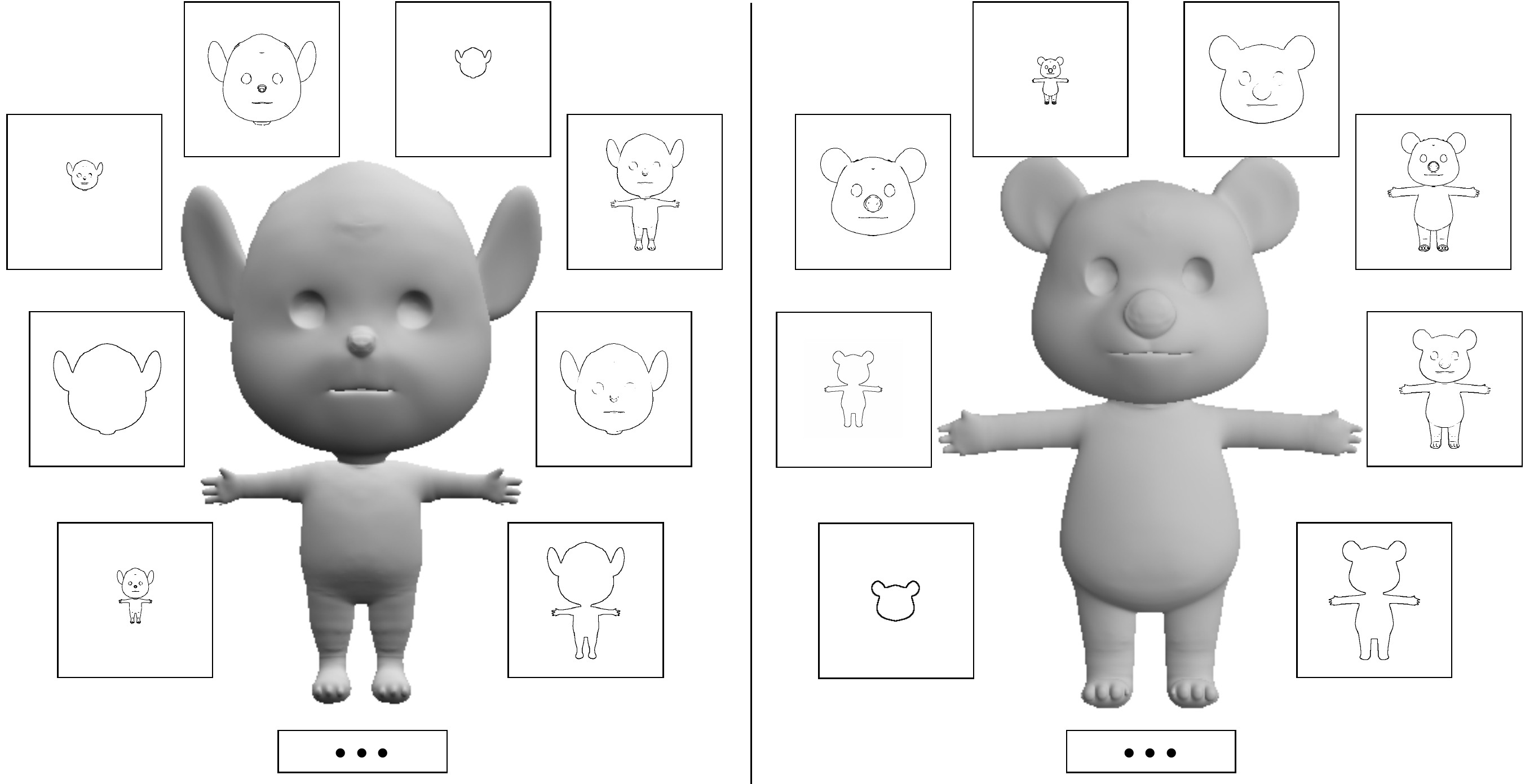}
  \caption{An illustration of rendered sketches used for training.}
  \label{fig_sketch_data}
\end{figure}

%% file: supp_figure/fig_sketch_pipeline.tex
\begin{figure}[htbp]
  \centering
  \includegraphics[width=.95 \linewidth]{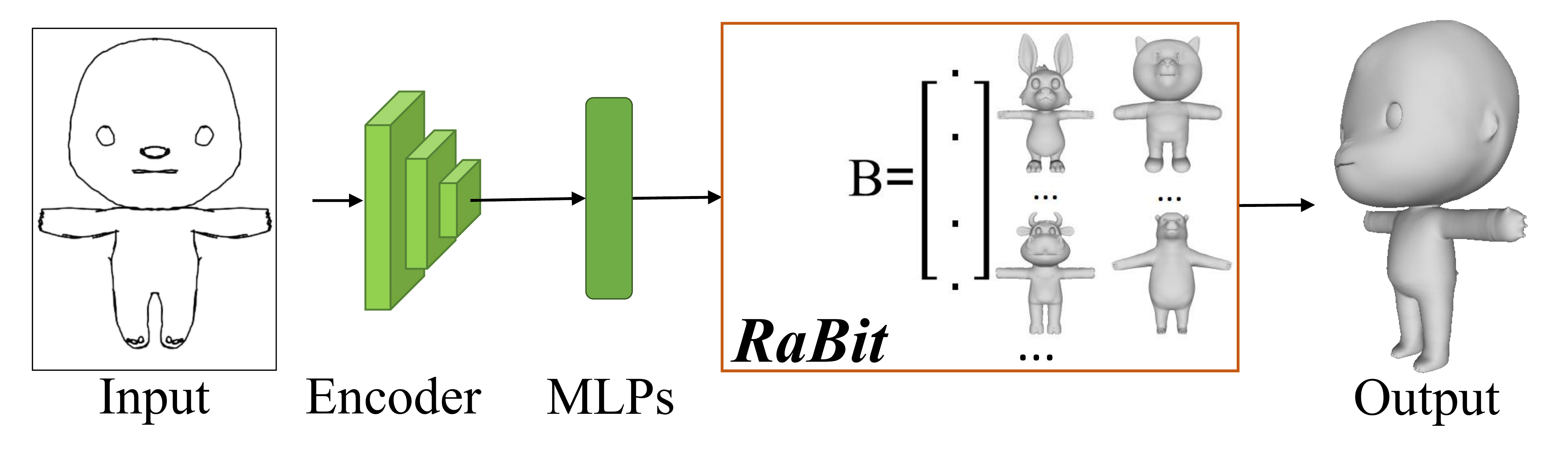}
  \caption{The pipeline of our sketch-based modeling. Given a sketch $512 \times 512$ as input, we employ one ResNet-50 module and three MLPs to embed the input to 100-dimensional shape parameters. The output shape parameters are fed to \textit{RaBit} to reconstruct the corresponding 3D model.}
  \label{fig_sketch_pipeline}
\end{figure}